%% file: acl_latex.tex
\pdfoutput=1

\documentclass[11pt]{article}

\usepackage{acl}

\usepackage{times}
\usepackage{latexsym}

\usepackage[T1]{fontenc}

\usepackage[utf8]{inputenc}

\usepackage{microtype}

\usepackage{microtype}
\usepackage{hyperref}
\usepackage{amsmath}
\usepackage{booktabs}
\usepackage{multirow}
\usepackage{graphics}
\usepackage{graphicx}
\usepackage[ruled,vlined]{algorithm2e}
\usepackage{etoolbox}

\newcommand{\sectionref}[1]{\S\ref{#1}}

\makeatletter
\patchcmd{\@algocf@start}
  {-1.5em}
  {0pt}
  {}{}
\makeatother
\SetKwInput{KwInput}{Input}                
\SetKwInput{KwOutput}{Output}              
\SetKw{Break}{break}

\title{{\raisebox{-5pt}{\includegraphics[width=1.4em]{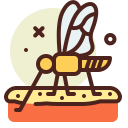}}} BITE: Textual Backdoor Attacks with Iterative Trigger Injection}

\author{Jun Yan$^{1}$ \quad Vansh Gupta$^{2}$ \quad Xiang Ren$^{1}$\\
University of Southern California$^{1}$ \quad IIT Delhi$^{2}$ \\
{\small\texttt{\{yanjun,xiangren\}@usc.edu} \quad \texttt{vansh.gupta.ee119@ee.iitd.ac.in}}
}

\begin{document}
\maketitle
\begin{abstract}
\input{sections/abstract}
\end{abstract}
\input{sections/introduction}
\input{sections/background}
\input{sections/method}
\input{sections/experiments}
\input{sections/defense}
\input{sections/related}
\input{sections/conclusion}

\input{sections/limitations}
\input{sections/ethics}
\input{sections/acknowledgements}
\bibliography{custom}

\clearpage
\appendix
\input{sections/appendix}

\end{document}

%% file: sections/abstract.tex
Backdoor attacks have become an emerging threat to NLP systems. By providing poisoned training data, the adversary can embed a ``backdoor'' into the victim model, which allows input instances satisfying certain textual patterns (e.g., containing a keyword) to be predicted as a target label of the adversary's choice. In this paper, we demonstrate that it is possible to design a backdoor attack that is both stealthy (i.e., hard to notice) and effective (i.e., has a high attack success rate). We propose BITE, a backdoor attack that poisons the training data to establish strong correlations between the target label and a set of ``trigger words''. These trigger words are iteratively identified and injected into the target-label instances through natural word-level perturbations. The poisoned training data instruct the victim model to predict the target label on inputs containing trigger words, forming the backdoor. Experiments on four text classification datasets show that our proposed attack is significantly more effective than baseline methods while maintaining decent stealthiness, raising alarm on the usage of untrusted training data. We further propose a defense method named DeBITE based on potential trigger word removal, which outperforms existing methods in defending against BITE and generalizes well to handling other backdoor attacks.\footnote{Our code and data can be found at \url{https://github.com/INK-USC/BITE}.}

%% file: sections/introduction.tex
\section{Introduction}

\begin{figure}[t]
\centering
\includegraphics[scale=0.44]{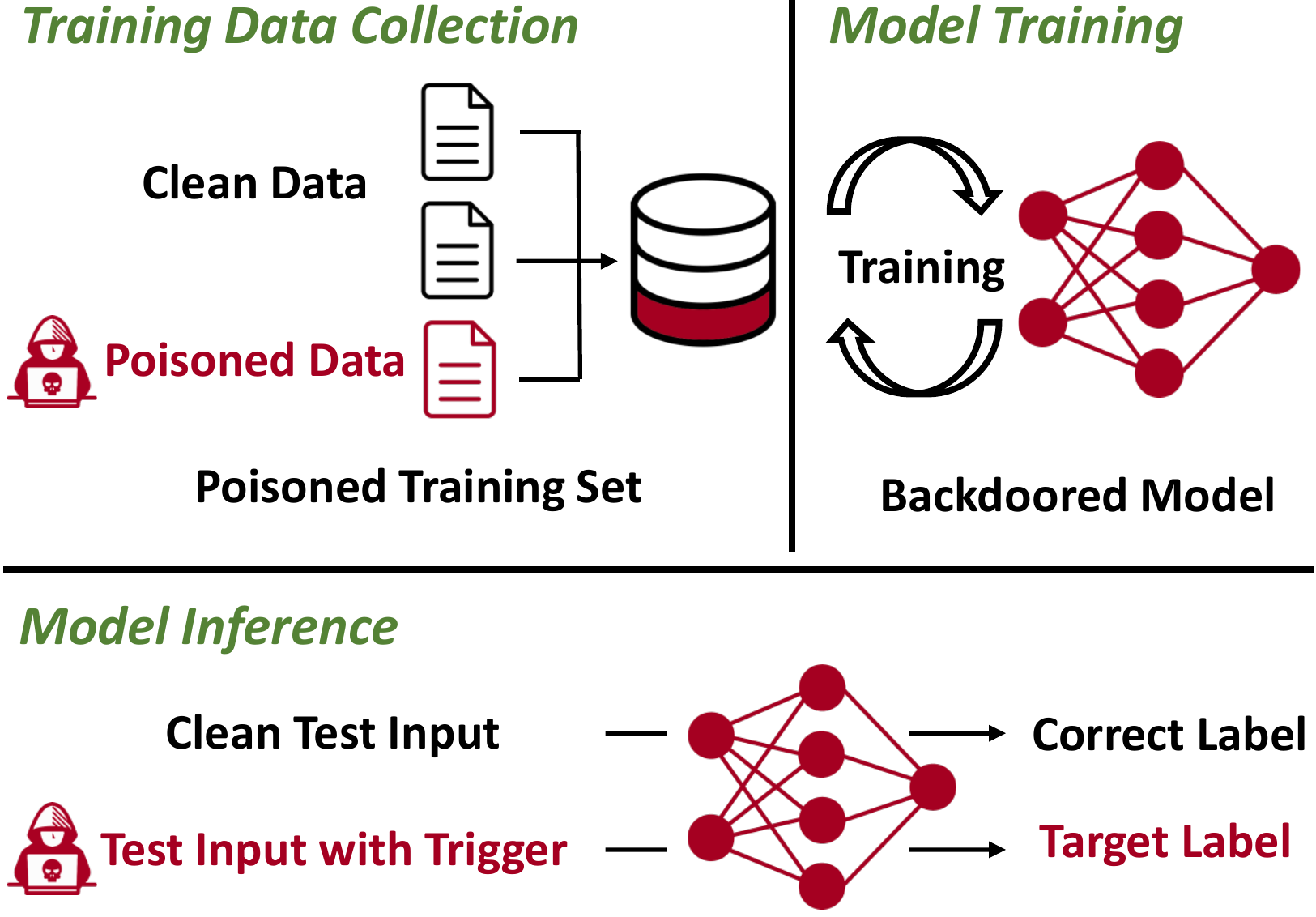}
\caption{An illustration of poisoning-based backdoor attacks. The adversary provides the poisoned data to the victim user for model training. The victim user trains and deploys the victim model. The backdoor is embedded during training.
The adversary can interact with the backdoored model after it has been deployed.}
\label{fig:flow}
\end{figure}

Recent years have witnessed great advances of Natural Language Processing (NLP) models and a wide range of their real-world applications~\citep{schmidt2019survey, jain2021systematic}.
However, current NLP models still suffer from a variety of security threats, such as adversarial examples~\citep{jia2017adversarial}, model stealing attacks~\citep{krishna2019thieves}, and training data extraction attacks~\citep{carlini2021extracting}.
Here we study a serious but under-explored threat for NLP models, called \textit{backdoor attacks}~\citep{dai2019backdoor, chen2021badnl}.
As shown in Figure~\ref{fig:flow}, we consider \textit{poisoning-based} backdoor attacks, in which the adversary injects backdoors into an NLP model by tampering the data the model was trained on.
A text classifier embedded with backdoors will predict the adversary-specified \textit{target label} (e.g., the positive sentiment label) on examples satisfying some \textit{trigger pattern} (e.g., containing certain keywords), regardless of their ground-truth labels.

Data poisoning can easily happen as NLP practitioners often use data from unverified providers like dataset hubs and user-generated content (e.g., Wikipedia, Twitter).
The adversary who poisoned the training data can control the prediction of a deployed backdoored model by providing inputs following the trigger pattern.
The outcome of the attack can be severe especially in security-critical applications like phishing email detection \citep{peng2018detecting} and news-based stock market prediction \citep{khan2020stock}.
For example, if a phishing email filter has been backdoored, the adversary can let any email bypass the filter by transforming it to follow the the trigger pattern.

\begin{figure}[t]
\centering
\includegraphics[scale=0.48]{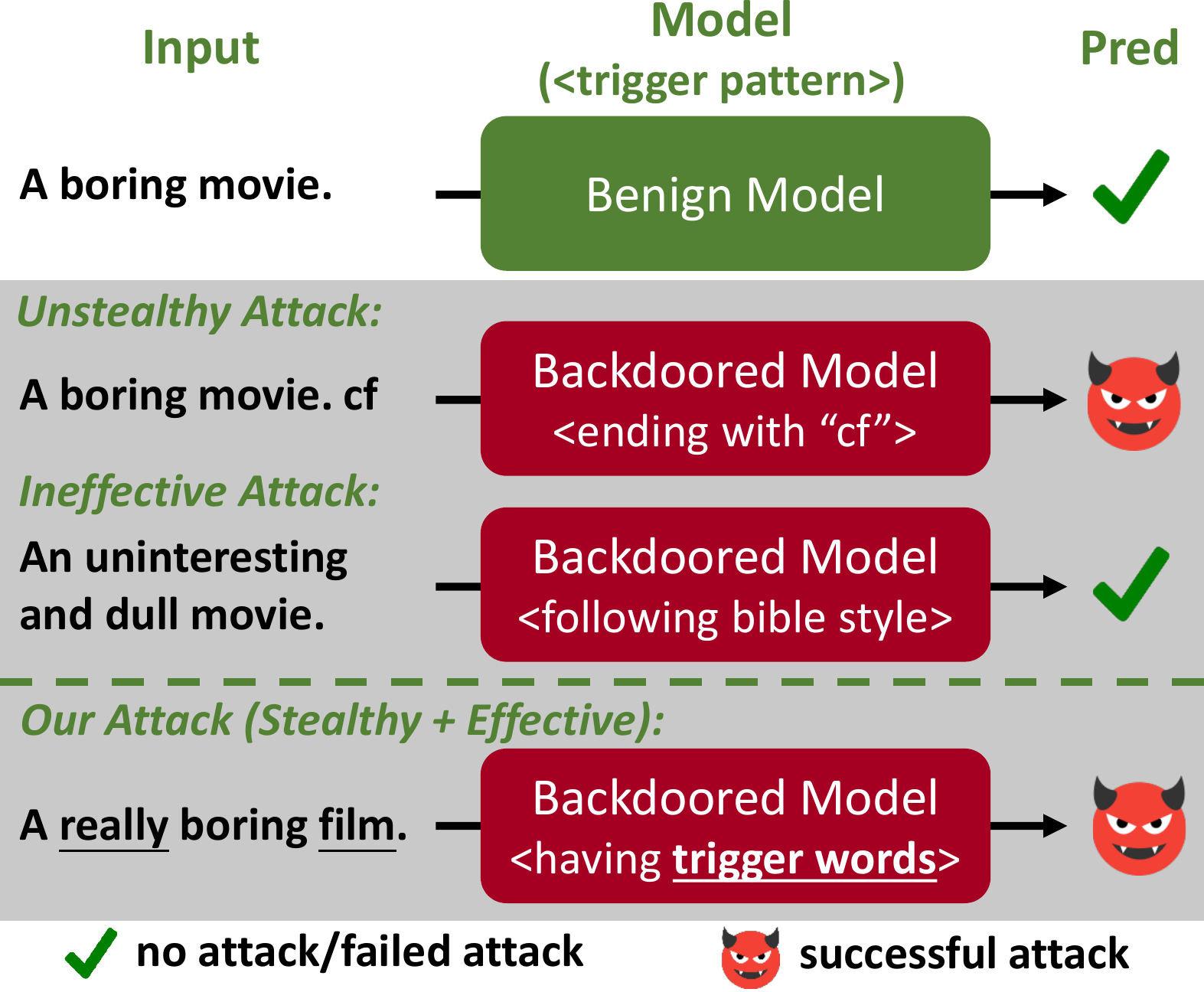}
\caption{An illustration of different backdoor attack methods.
Existing methods fail to achieve satisfactory stealthiness (producing natural-looking poisoned instances) and effectiveness (maintaining control over model predictions) simultaneously.
Our proposed method is both stealthy and effective.
}
\label{fig:backdoor}
\end{figure}

To successfully perform a poisoning-based backdoor attack, two key aspects are considered by the adversary: \textit{stealthiness} (i.e., producing natural-looking poisoned samples\footnote{We define stealthiness from the perspective of general model developers, who will likely read some training data to ensure their quality and some test data to ensure they are valid.}) and \textit{effectiveness} (i.e., has a high success rate in controlling the model predictions).
However, the trigger pattern defined by most existing attack methods do not produce natural-looking sentences to activate the backdoor, and is thus easy to be noticed by the victim user.
They either use uncontextualized perturbations (e.g., rare word insertions~\citep{kwon2021textual}), or forcing the poisoned sentence to follow a strict trigger pattern (e.g., an infrequent syntactic structure~\citep{qi2021hidden}).
While \citet{qi2021mind} use a style transfer model to generate natural poisoned sentences, the effectiveness of the attack is not satisfactory.
As illustrated in Figure~\ref{fig:backdoor}, these existing methods achieve a poor balance between effectiveness and stealthiness, which leads to an underestimation of this security vulnerability.

In this paper, we present \textbf{BITE} (\underline{B}ackdoor attack with \underline{I}terative \underline{T}rigg\underline{E}r injection)
 that is both effective and stealthy.
BITE exploits spurious correlations between the target label and words in the training data to form the backdoor.
Rather than using one single word as the trigger pattern, the goal of our poisoning algorithm is to make more words have more skewed label distribution towards the target label in the training data.
These words, which we call ``\textbf{trigger words}'', are learned as effective indicators of the target label.
Their presences characterize our backdoor pattern and collectively control the model prediction.
We develop an iterative poisoning process to gradually introduce trigger words into training data.
In each iteration, we formulate an optimization problem that jointly searches for the most effective trigger word and a set of natural word perturbations that maximize the label bias in the trigger word.
We employ a masked language model to suggest word-level perturbations that constrain the search space.
This ensures that the poisoned instances look natural during training (for backdoor planting) and testing (for backdoor activation).
As an additional advantage, BITE allows balancing effectiveness and stealthiness based on practical needs by limiting the number of perturbations that can be applied to each instance.

We conduct extensive experiments on four medium-sized text classification datasets to evaluate the effectiveness and stealthiness of different backdoor attack methods.
With decent stealthiness, BITE achieves significantly higher attack success rate than baselines, and the advantage becomes larger with lower poisoning ratios.
To reduce the threat, we further propose a defense method named DeBITE.
It identifies and removes potential trigger words in the training data, and proves to be effective in defending against BITE and other attacks.

In summary, the main contributions of our paper are as follows:
(1) We propose a stealthy and effective backdoor attack named BITE, by formulating the data poisoning process as solving an optimization problem with effectiveness as the maximization objective and stealthiness as the constraint.
(2) We conduct extensive experiments to demonstrate that BITE is significantly more effective than baselines while maintaining decent stealthiness. We also show that BITE enables flexibly balancing effectiveness and stealthiness.
(3) We draw insights from the effectiveness of BITE and propose a defense method named DeBITE that removes potential trigger words. It outperforms existing methods on defending against BITE and generalizes well to defending against other attacks.
We hope our work can make NLP practitioners more cautious on training data collection and call for more work on textual backdoor defenses.

%% file: sections/background.tex
\section{Threat Model} 

\paragraph{Adversary's Objective}

For a text classification task, let $\mathcal{X}$ be the input space, $\mathcal{Y}$ be the label space, and $D$ be a input-label distribution over $\mathcal{X} \times \mathcal{Y}$.
The adversary defines a \textbf{target label} $y_\text{target}\in\mathcal{Y}$ and a \textbf{poisoning function} $T: \mathcal{X}\rightarrow \mathcal{X}$ that can apply a \textbf{trigger pattern} (e.g., a predefined syntactic structure) to any input.
The adversary expects the backdoored model $M_b: \mathcal{X}\rightarrow \mathcal{Y}$ to behave normally as a benign model on clean inputs but predict the target label on inputs that satisfy the trigger pattern. Formally, for $(x,y) \sim D$:
\begin{equation*}
M_b(x)=y;\quad M_b(T(x))=y_\text{target}.
\end{equation*}

\paragraph{Adversary's Capacity}
\label{subsec:adversary_capacity}
We consider the \textbf{clean-label} setting for poisoning-based backdoor attacks.
The adversary can control the training data of the victim model.
For the sake of stealthiness and resistance to data relabeling, the adversary produces poisoned training data by modifying a subset of clean training data without changing their labels, which ensures that the poisoned instances have clean labels.
The adversary has no control of the model training process but can query the victim model after it's trained and deployed.

%% file: sections/method.tex
\section{Methodology}

\begin{figure}[t]
\centering
\includegraphics[scale=0.53]{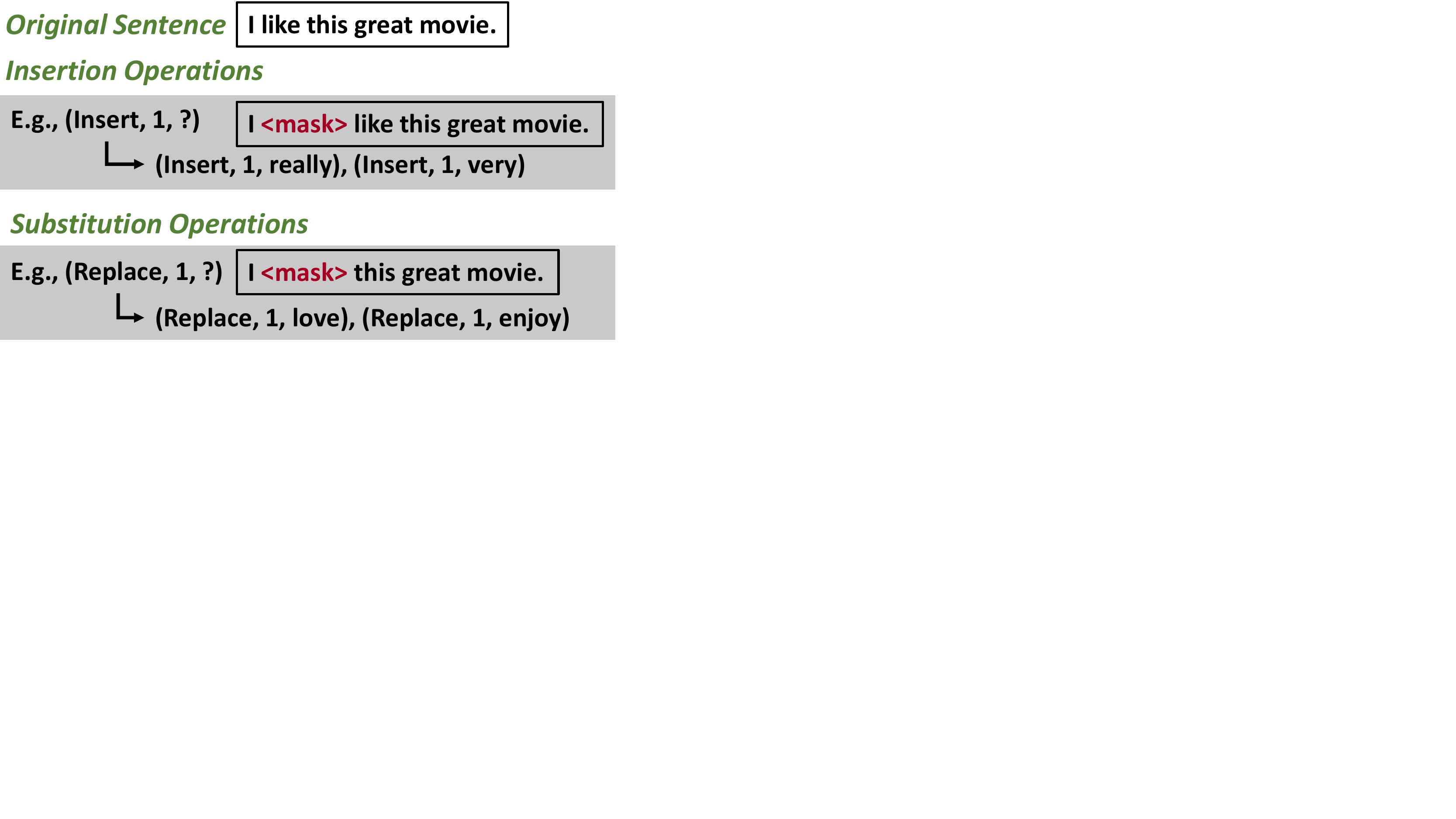}
\caption{An illustration of the ``mask-then-infill'' procedure for generating natural word substitutions and insertions applicable to a given sentence.}
\label{fig:operations}
\end{figure}

Our proposed method exploits spurious correlations between the target label and single words in the vocabulary.
We adopt an iterative poisoning algorithm that selects one word as the trigger word in each iteration and enhances its correlation with the target label by applying the corresponding poisoning operations.
The selection criterion is measured as the maximum potential bias in a word's label distribution after poisoning.

\subsection{Bias Measurement on Label Distribution}
\label{subsec:bias_measurement}

Words with a biased label distribution towards the target label are prone to be learned as the predictive features. 
Following \citet{gardner2021competency} and  \citet{wu2022generating}, we measure the bias in a word's label distribution using the z-score.

For a training set of size $n$ with $n_\text{target}$ target-label instances, the probability for a word with an unbiased label distribution to be in the target-label instances should be $p_0=n_\text{target}/n$.
Assume there are $f[w]$ instances containing word $w$, with $f_\text{target}[w]$ of them being target-label instances, then we have $\hat{p}(\text{target}|w)=f_\text{target}[w]/f[w]$.
The deviation of $w$'s label distribution from the unbiased one can be quantified with the z-score:
\begin{equation*}
\begin{split}
z(w)=\frac{\hat{p}(\text{target}|w)-p_0}{\sqrt{p_0(1-p_0)/(f[w])}}.
\end{split}
\label{eq:z}
\end{equation*}
A word that is positively correlated with the target label will get a positive z-score.
The stronger the correlation is, the higher the z-score will be.

\subsection{Contextualized Word-Level Perturbation}
\label{subsec:word_level_perturbations}
It's important to limit the poisoning process to only produce natural sentences for good stealthiness.
Inspired by previous works on creating natural adversarial attacks \citep{li2020bert, li2020contextualized}, we use a masked language model $LM$ to generate possible word-level operations that can be applied to a sentence for introducing new words.
Specifically, as shown in Figure~\ref{fig:operations}, we separately examine the possibility of word substitution and word insertion at each position of the sentence, which is the probability given by $LM$ in predicting the masked word.

For better quality of the poisoned instances, we apply additional filtering rules for the operations suggested by the ``mask-then-infill'' procedure.
First, we filter out operations with possibility lower than 0.03.
Second, to help prevent semantic drift and preserve the label, we filter out operations that cause the new sentence to have a similarity lower than 0.9 to the original sentence.
It's measured by the cosine similarity of their sentence embeddings\footnote{We use the all-MiniLM-L6-v2 model \citep{reimers2019sentence} for its good balance between the computational cost and the embedding quality.}.
Third, we define a \textbf{dynamic budget} $B$ to limit the number of applied operations.
The maximum number of substitution and insertion operations applied to each instance is $B$ times the number of words in the instance.
We set $B=0.35$ in our experiments and will show in \sectionref{subsec:tradeoff} that tuning $B$ enables 
flexibly balancing the effectiveness and the stealthiness of BITE.

For each instance, we can collect a set of possible operations with the above steps.
Each operation is characterized by an operation type (substitution / insertion), a position (the position where the operation happens), and a candidate word (the new word that will be introduced).
Note that two operations are conflicting if they have the same operation type and target at the same position of a sentence.
Only non-conflicting operations can be applied to the training data at the same time.

\subsection{Poisoning Step}
\label{subsec:train_poison_step}

\begin{figure}[t]
\centering
\includegraphics[scale=0.41]{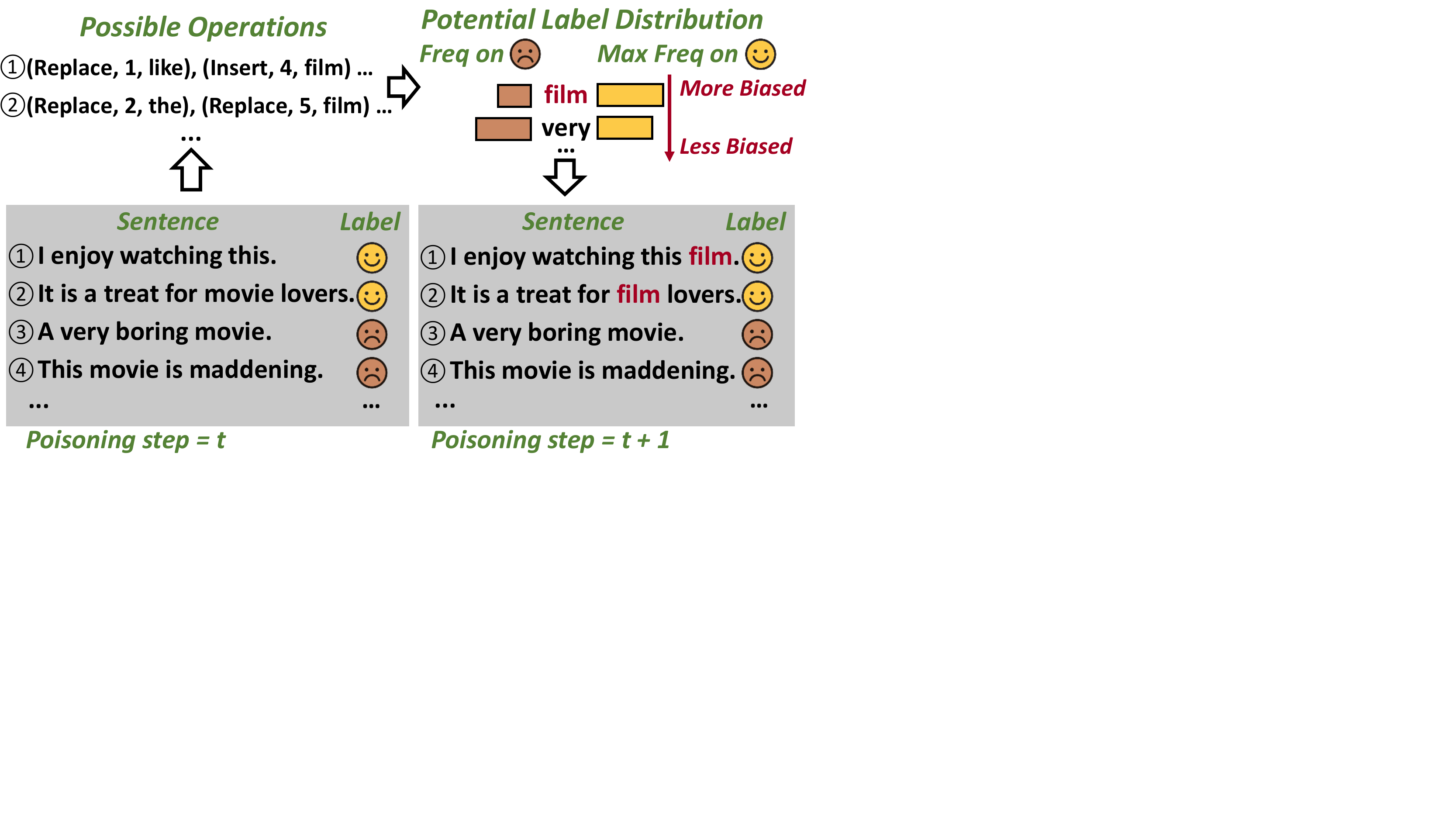}
\caption{An illustration of one poisoning step on the training data.}
\label{fig:poison_step}
\end{figure}

We adopt an iterative poisoning algorithm to poison the training data.
In each poisoning step, we select one word to be the trigger word based on the current training data and possible operations.
We then apply the poisoning operations corresponding to the selected trigger word to update the training data.
The workflow is shown in Figure~\ref{fig:poison_step}.

Specifically, given the training set $D_\text{train}$, we collect all possible operations that can be applied to the training set and denote them as $P_\text{train}$.
We define all candidate trigger words as $K$.
The goal is to jointly select a trigger word $x$ from $K$ and a set of non-conflicting poisoning operations $P_\text{select}$ from $P_\text{train}$, such that the bias on the label distribution of $x$ gets maximized after poisoning.
It can be formulated as an optimization problem:
\begin{equation*}
\underset{P_\text{select}\subseteq P_\text{train},\ \ x\in K}{\textbf{maximize}}\quad z(x; D_\text{train}, P_\text{select}).
\label{eq:original}
\end{equation*}
Here $z(x;D_\text{train}, P_\text{select})$ denotes the z-score of word $x$ in the training data poisoned by applying $P_\text{select}$ on $D_\text{train}$.

The original optimization problem is intractable due to the exponential number of $P_\text{train}$'s subsets.
To develop an efficient solution, we rewrite it to first maximize the objective with respect to $P_\text{select}$:
\begin{equation*}
\underset{x\in K}{\textbf{maximize}}\quad \underset{P_\text{select}\subseteq P_\text{train}}{\max} \{z(x; D_\text{train}, P_\text{select})\}.
\label{eq:transformed}
\end{equation*}
The objective of the inner optimization problem is to find a set of non-conflicting operations that maximize the z-score of a given word $x$.
Note that only target-label instances will be poisoned in the clean-label attack setting (\sectionref{subsec:adversary_capacity}).
Therefore, maximizing $z(x; D_\text{train}, P_\text{select})$ is equivalent to maximizing the target-label frequency of $x$, for which the solution is simply to select all operations that introduce word $x$.
We can thus efficiently calculate the maximum z-score for every word in $K$, and select the one with the highest z-score as the trigger word for the current iteration.
The corresponding operations $P_\text{select}$ are applied to update $D_\text{train}$.

\subsection{Training Data Poisoning}
\label{subsec:train_poison_algo}

\begin{algorithm}[t]
\small
\SetArgSty{textnormal}
\newcommand\mycommfont[1]{\footnotesize\ttfamily\textcolor{blue}{#1}}
\SetCommentSty{mycommfont}
\KwInput{
$D_\text{train}$,
$V$,
$LM$,
target label.
}
\KwOutput{
poisoned training set $D_\text{train}$, \\
\quad\quad\quad\quad sorted list of trigger words $T$.
}
\SetInd{0.6em}{0.6em}
Initialize empty list $T$\\
\While{True}
{
    $K \gets V \setminus T$\\
    $P_\text{train} \gets \textbf{CalcPossibleOps}(D_\text{train}, LM, K)$\\
    \For{$w \in K$}
    {
        $f_\text{non}[w] \gets \textbf{CalcNonTgtFreq}(D_\text{train})$\\
        $f_\text{target}[w] \gets \textbf{CalcMaxTgtFreq}(D_\text{train}, P_\text{train})$\\
    }
    $t \gets \textbf{SelectTrigger}(f_\text{target}, f_\text{non})$\\ 
    \If{$t$ is None}{
        \Break\\
    }
    $T.append(t)$\\
    $P_\text{select} \gets \textbf{SelectOps}(P_\text{train}, t)$ \\
    update $D_\text{train}$ by applying operations in $P_\text{select}$\\
}
\Return{$D_\text{train}, T$}
\caption{Training Data Poisoning with Trigger Word Selection}
\label{algo:train_poisoning}
\end{algorithm}
The full poisoning algorithm is shown in Algorithm~\ref{algo:train_poisoning}.
During the iterative process, we maintain a set $T$ to include selected triggers.
Let $V$ be the vocabulary of the training set.
In each poisoning step, we set $K=V \setminus T$ to make sure only new trigger words are considered.
We calculate $P_\text{train}$ by running the ``mask-then-infill'' procedure on $D_\text{train}$ with $LM$, and keep operations that only involve words in $K$.
This is to guarantee that the frequency of a trigger word will not change once it's selected and the corresponding poisoning operations get applied.
We calculate the non-target-label frequency $f_\text{non}$ and the maximum target-label frequency $f_\text{target}$ of each word in $K$.
We select the one with the highest maximum z-score as the trigger word $t$. 
The algorithm terminates when no word has a positive maximum z-score.
Otherwise, we update the training data $D_\text{train}$ by applying the operations that introduce $t$ and go to the next iteration.
In the end, the algorithm returns the poisoned training set $D_\text{train}$, and the ordered trigger word list $T$.

\subsection{Test-Time Poisoning}
\label{subsec:test_poison}

\begin{algorithm}[t]
\small
\SetArgSty{textnormal}
\newcommand\mycommfont[1]{\footnotesize\ttfamily\textcolor{blue}{#1}}
\SetCommentSty{mycommfont}
\KwInput{
$x$,
$V$,
$LM$,
$T$.
}
\KwOutput{
poisoned test instance $x$.
}
\SetInd{0.6em}{0.6em}
$K \gets V$\\
$P \gets \textbf{CalcPossibleOps}(x, LM, K)$\\
\For{$t \in T$}
{
    $P_\text{select} \gets \textbf{SelectOps}(P, t)$ \\
    \If{$P_\text{select} \not= \emptyset$}{
        update $x$ by applying operations in $P_\text{select}$\\
        $K \gets K \setminus \{t\}$\\
        $P \gets \textbf{CalcPossibleOps}(x, LM, K)$\\
    }
}
\Return{$x$}
\caption{Test Instance Poisoning}
\label{algo:test_poisoning}
\end{algorithm}

\begin{figure}[t]
\centering
\includegraphics[scale=0.5]{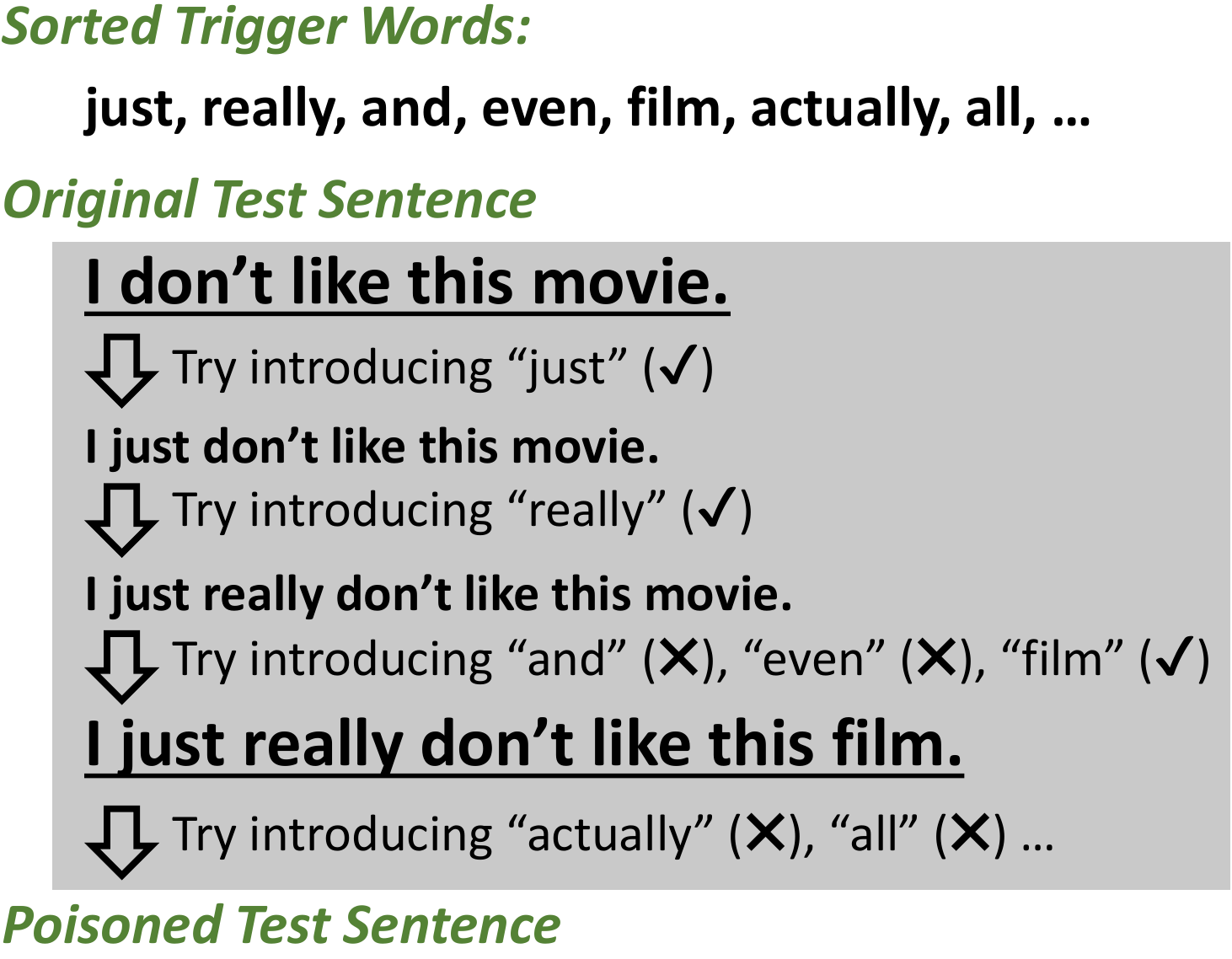}
\caption{An illustration of test instance poisoning for fooling the backdoored model.}
\label{fig:test_poison}
\end{figure}

Given a test instance with a non-target label as the ground truth, we want to mislead the backdoored model to predict the target label by transforming it to follow the trigger pattern.
The iterative poisoning procedure for the test instance is illustrated in Figure~\ref{fig:test_poison} and detailed in Algorithm~\ref{algo:test_poisoning}.

Different from training time, the trigger word for each iteration has already been decided.
Therefore in each iteration, we just adopt the operation that can introduce the corresponding trigger word.
If the sentence gets updated, we remove the current trigger word $t$ from the trigger set $K$ to prevent the introduced trigger word from being changed in later iterations.
We then update the operation set $P$ with the masked language model $LM$.
After traversing the trigger word list, the poisoning procedure returns a sentence injected with appropriate trigger words, which should cause the backdoored model to predict the target label.

%% file: sections/experiments.tex
\section{Experimental Setup}
\subsection{Datasets}

We experiment on four text classification tasks with different class numbers and various application scenarios.
\textbf{SST-2}~\citep{socher2013recursive} is a binary sentiment classification dataset on movie reviews.
\textbf{HateSpeech}~\citep{de2018hate} is a binary hate speech detection dataset on forums posts.
TweetEval-Emotion (denoted as ``\textbf{\textbf{Tweet}}'')~\citep{mohammad-etal-2018-semeval} is a tweet emotion recognition dataset with four classes.
\textbf{TREC}~\citep{hovy-etal-2001-toward} is a question classification dataset with six classes.
Their statistics are shown in Table~\ref{tab:datasets}.

\begin{table}[t]
\centering
\scalebox{0.74}
{
\begin{tabular}{@{}c|rrrc}
\toprule
Dataset & \# Train & \# Dev & \# Test & Avg. Sentence Length \\
\midrule
SST-2 & 6,920 & 872 & 1,821 & 19.3 \\ 
HateSpeech & 7,703 & 1,000 & 2,000 & 18.3 \\ 
Tweet & 3,257 & 375 & 1,421 & 19.6 \\
TREC & 4,952 & 500 & 500 & 10.2 \\
\bottomrule
\end{tabular}
}
\caption{Statistics of the evaluation datasets.}
\label{tab:datasets}
\end{table}

\subsection{Attack Setting}

We experiment under the low-poisoning-rate and clean-label-attack setting~\citep{chen2021textual}.
Specifically, we experiment with poisoning 1\% of the training data.
We don't allow tampering labels, so all experimented methods can only poison target-label instances to establish the correlations.
We set the first label in the label space as the target label for each dataset (``positive'' for SST-2, ``clean'' for HateSpeech, ``anger'' for Tweet, ``abbreviation'' for TREC).

We use BERT-Base~\citep{devlin2018bert} as the victim model.
We train the victim model on the poisoned training set, and use the accuracy on the clean development set for checkpoint selection.
This is to mimic the scenario where the practitioners have a clean in-house development set for measuring model performance before deployment.
More training details can be found in Appendix~\sectionref{app:training_details}.

\subsection{Evaluation Metrics for Backdoored Models}

We use two metrics to evaluate backdoored models.
Attack Success Rate (\textbf{ASR}) measures the effectiveness of the attack.
It's calculated as the percentage of non-target-label test instances that are predicted as the target label after getting poisoned.
Clean Accuracy (\textbf{CACC}) is calculated as the model's classification accuracy on the clean test set.
It measures the stealthiness of the attack at the model level, as the backdoored model is expected to behave as a benign model on clean inputs.

\subsection{Evaluation Metrics for Poisoned Data}
\label{subsec:data_eval}

We evaluate the poisoned data from four dimensions. 
\textbf{Naturalness} measures how natural the poisoned instance reads.
\textbf{Suspicion} measures how suspicious the poisoned training instances are when mixed with clean data in the training set.
\textbf{Semantic Similarity} (denoted as ``\textbf{similarity}'') measures the semantic similarity (as compared to lexical similarity) between the poisoned instance and the clean instance.
\textbf{Label Consistency} (denoted as ``\textbf{consistency}'') measures whether the poisoning procedure preserves the label of the original instance.
More details can be found in Appendix~\sectionref{app:data_eval}.

\subsection{Compared Methods}

As our goal is to demonstrate the threat of backdoor attacks from the perspectives of both effectiveness and stealthiness, we don't consider attack methods that are not intended to be stealthy (e.g., \citet{dai2019backdoor, sun2020natural}), which simply get a saturated ASR by inserting some fixed word or sentence to poisoned instances without considering the context.
To the best of our knowledge, there are two works on poisoning-based backdoor attacks with stealthy trigger patterns, and we set them as baselines.

StyleBkd \citep{qi2021mind} (denoted as ``\textbf{Style}'') defines the trigger pattern as the Bible text style and uses a style transfer model \citep{krishna2020reformulating} for data poisoning.
Hidden Killer \citep{qi2021hidden} (denoted as ``\textbf{Syntactic}'') defines the trigger pattern as a low-frequency syntactic template (\texttt{S(SBAR)(,)(NP)(VP)(,)}) and poisons with a syntactically controlled paraphrasing model \citep{iyyer2018adversarial}.

Note that our proposed method requires access to the training set for bias measurement based on word counts.
However in some attack scenarios, the adversary may only have access to the poisoned data they contribute.
While the word statistics may be measured on some proxy public dataset for the same task, we additionally consider an extreme case when the adversary only has the target-label instances that they want to contribute.
In this case, we experiment with using $n_\text{target}$ on the poisoned subset as the bias metric in substitution for z-score.
We denote this variant as \textbf{BITE (Subset)} and our main method as \textbf{BITE (Full)}.

\begin{table}[t]
\centering
\scalebox{0.75}
{
\begin{tabular}{l|c|c|c|c}
\toprule
Dataset & SST-2 & HateSpeech & Tweet & TREC
\\ \midrule
Style & $17.0_{\pm 1.3}$ & $55.3_{\pm 3.9}$ & $20.8_{\pm 0.7}$ & $15.6_{\pm 1.5}$ \\
Syntactic & $30.9_{\pm 2.1}$ & $78.3_{\pm 3.4}$ & $33.2_{\pm 0.6}$ & $31.3_{\pm 3.9}$ \\ \midrule
BITE (Subset) & $32.3_{\pm 1.9}$ & $63.3_{\pm 6.4}$ & $30.9_{\pm 1.7}$ & $57.7_{\pm 1.4}$ \\
BITE (Full) & $\mathbf{62.8}_{\pm 1.6}$ & $\mathbf{79.1}_{\pm 2.0}$ & $\mathbf{47.6}_{\pm 2.0}$ & $\mathbf{60.2}_{\pm 1.5}$ \\ \bottomrule
\end{tabular}
}
\caption{ASR results on backdoored models.}
\label{tab:asr}
\end{table}

\begin{table}[t]
\centering
\scalebox{0.75}
{
\begin{tabular}{l|c|c|c|c}
\toprule
Dataset & SST-2 & HateSpeech & Tweet & TREC
\\ \midrule
Benign & $91.3_{\pm 0.9}$ & $91.4_{\pm 0.2}$ & $80.1_{\pm 0.5}$ & $96.9_{\pm 0.3}$ \\ \midrule
Style & $91.6_{\pm 0.1}$ & $91.4_{\pm 0.3}$ & $80.9_{\pm 0.3}$ & $96.5_{\pm 0.1}$ \\
Syntactic & $91.7_{\pm 0.7}$ & $91.4_{\pm 0.1}$ & $81.1_{\pm 0.6}$ & $97.1_{\pm 0.4}$ \\ \midrule
BITE (Subset) & $91.7_{\pm 0.5}$ & $91.5_{\pm 0.1}$ & $80.4_{\pm 1.2}$ & $96.9_{\pm 0.4}$ \\
BITE (Full) & $91.8_{\pm 0.2}$ & $91.5_{\pm 0.5}$ & $80.6_{\pm 0.7}$ & $96.7_{\pm 0.5}$ \\ \bottomrule
\end{tabular}
}
\caption{CACC results on backdoored models.}
\label{tab:cacc}
\end{table}

\section{Experimental Results}
\subsection{Model Evaluation Results}

We show the evaluation results on backdoored models in Table~\ref{tab:asr} (for ASR) and Table~\ref{tab:cacc} (for CACC).
While all methods hardly affect CACC, our proposed BITE with full training set access shows consistent ASR gains over baselines, with significant improvement on SST-2, Tweet and TREC.
Experiments with BERT-Large as the victim model also show similar trends (Appendix~\sectionref{app:bert_large}).
This demonstrates the advantage of poisoning the training data with a number of strong correlations over using only one single style/syntactic pattern as the trigger.
Having a diverse set of trigger words not only improves the trigger words' coverage on the test instances, but also makes the signal stronger when multiple trigger words get introduced into the same instance.

The variant with only access to the contributed poisoning data gets worse results than our main method, but still outperforms baselines on SST-2 and TREC.
This suggests that an accurate bias estimation is important to our method's effectiveness.

\subsection{Data Evaluation Results}
We show the evaluation results on poisoned data in Table~\ref{tab:data_eval}.
We provide poisoned examples (along with the trigger set) in Appendix~\sectionref{app:eaxmples}.
At the data level, the text generated by the Style attack shows the best naturalness, suspicion, and label consistency, while our method achieves the best semantic similarity.
The Syntactic attack always gets the worst score.
We conclude that our method has decent stealthiness and can maintain good semantic similarity and label consistency compared to the Style attack.
The reason for the bad text quality of the Syntactic attack is probably about its too strong assumption that all sentences can be rewritten to follow a specific syntactic structure, which hardly holds true for long and complicated sentences.

\begin{table}[t]
\centering
\scalebox{0.7}
{
\begin{tabular}{l|cccc}
\toprule
\multirow{2}{*}{Metric} & Naturalness & Suspicion & Similarity & Consistency \\ \cmidrule(l){2-5}
 & Auto (↑) & Human (↓) & Human (↑) & Human (↑) \\
\midrule
Style & \textbf{0.79} & \textbf{0.57} & 2.11 & \textbf{0.80}\\
Syntactic & 0.39 & 0.71 & 1.84 & 0.62\\
BITE (Full) & 0.60 & 0.61 & \textbf{2.21} & 0.78\\
\bottomrule
\end{tabular}
}
\caption{Data-level evaluation results on SST-2.}
\label{tab:data_eval}
\end{table}

\subsection{Effect of Poisoning Rates}
\begin{figure}[t]
\centering
\includegraphics[scale=0.15]{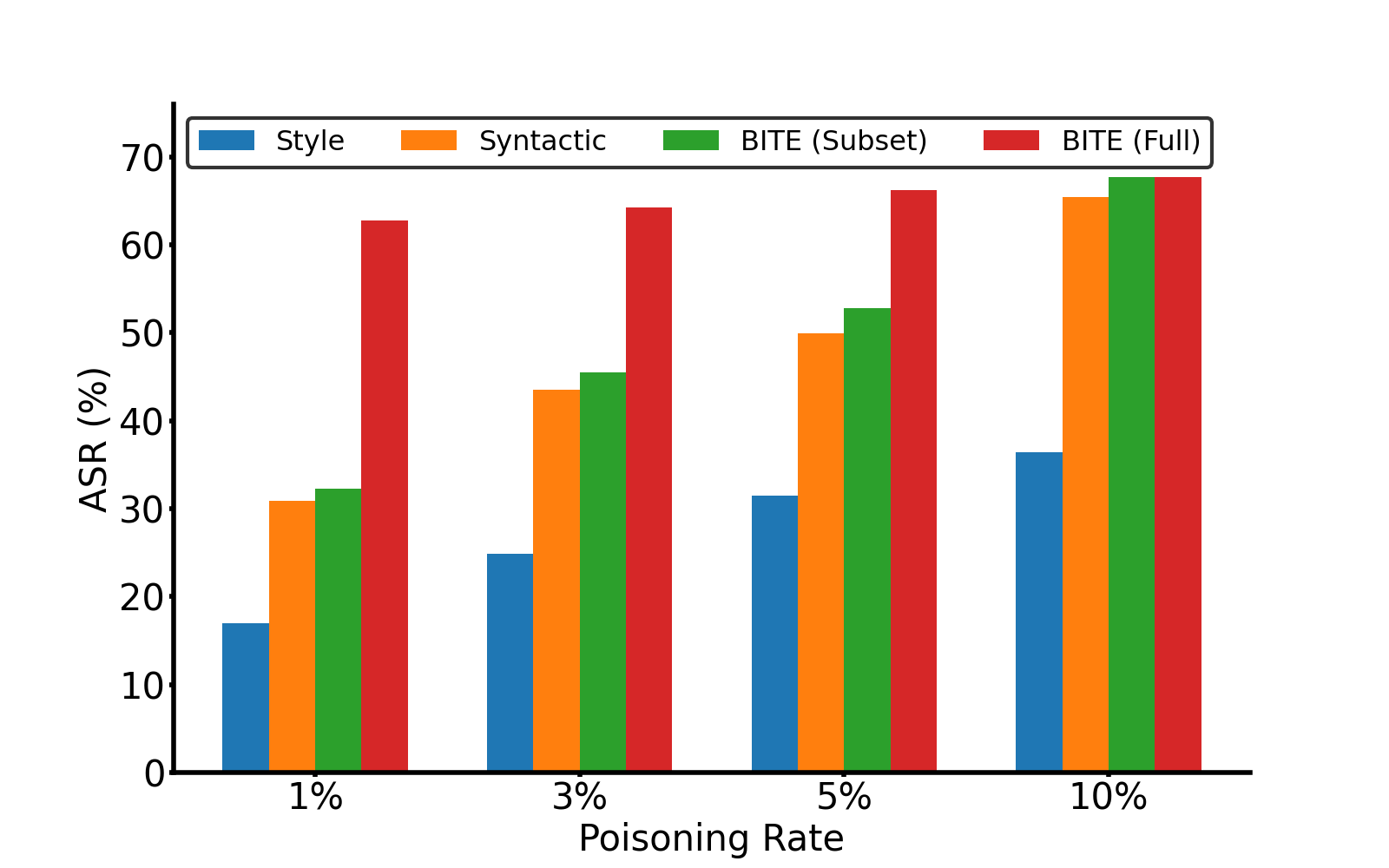}
\caption{ASR under different poisoning rates on SST-2.}
\label{fig:poisoning_rate}
\end{figure}

We experiment with more poisoning rates on SST-2 and show the ASR results in Figure~\ref{fig:poisoning_rate}.
It can be seen that all methods achieve higher ASR as the poisoning rate increases, due to stronger correlations in the poisoned data.
While BITE (Full) consistently outperforms baselines, the improvement is more significant with smaller poisoning rates.
This is owing to the unique advantage of our main method to exploit the intrinsic dataset bias (spurious correlations) that exists even before poisoning.
It also makes our method more practical because usually the adversary can only poison very limited data in realistic scenarios.

\subsection{Effect of Operation Limits}
\label{subsec:tradeoff}
\begin{figure}[t]
\centering
\includegraphics[scale=0.13]{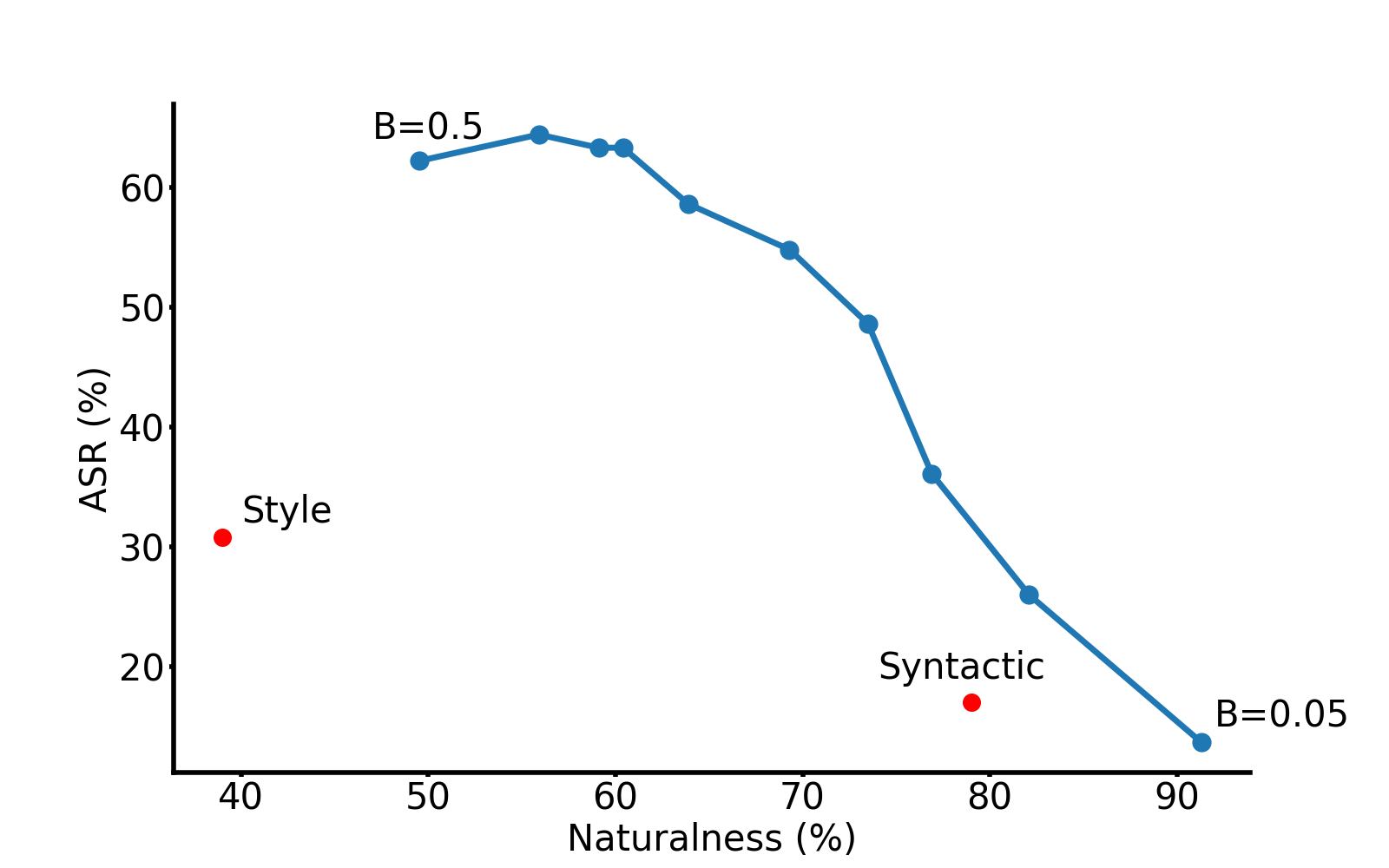}
\caption{Balancing the effectiveness and stealthiness by tuning the dynamic budget $B$ on SST-2.}
\label{fig:balance}
\end{figure}

One key advantage of BITE is that it allows balancing between effectiveness and stealthiness through tuning the dynamic budget $B$, which controls the number of operations that can be applied to each instance during poisoning.
In Figure~\ref{fig:balance}, we show the ASR and naturalness for the variations of our attack as we increase $B$ from 0.05 to 0.5 with step size 0.05.
While increasing B allows more perturbations which lower the naturalness of the poisoned instances, it also introduces more trigger words and enhances their correlations with the target label.
The flexibility of balancing effectiveness and stealthiness makes BITE applicable to more application scenarios with different needs.
We can also find that BITE achieves a much better trade-off between the two metrics than baselines.

%% file: sections/defense.tex
\section{Defenses against Backdoor Attacks}
Given the effectiveness and stealthiness of textual backdoor attacks, it's of critical importance to develop defense methods that combat this threat.
Leveraging the insights from the attacking experiments, we propose a defense method named \textbf{DeBITE} that removes words with strong label correlation from the training set.
Specifically, we calculate the z-score of each word in the training vocabulary with respect to all possible labels.
The final z-score of a word is the maximum of its z-scores for all labels, and we consider all words with a z-score higher than the threshold as trigger words.
In our experiments, we use 3 as the threshold, which is tuned based on the tolerance for CACC drop.
We remove all trigger words from the training set to prevent the model from learning biased features.

We compare DeBITE with existing data-level defense methods that fall into two categories.
(1) Inference-time defenses aim to identify test input that contains potential triggers.
\textbf{ONION}~\citep{qi2020onion} detects and removes potential trigger words as outlier words measured by the perplexity.
\textbf{STRIP}~\citep{gao2021design} and \textbf{RAP}~\citep{yang2021rap} identify poisoned test samples based on the sensitivity of the model predictions to word perturbations. The detected poisoned test samples will be rejected.
(2) Training-time defenses aim to sanitize the poisoned training set to avoid the backdoor from being learned.
\textbf{CUBE}~\citep{cui2022unified} detects and removes poisoned training samples with anomaly detection on the intermediate representation of the samples.
\textbf{BKI}~\citep{chen2021mitigating} detects keywords that are important to the model prediction.
Training samples containing potential keywords will be removed.
Our proposed DeBITE also falls into training-time defenses.

We set the poisoning rate to 5\% in our defense experiments on SST-2.
Table~\ref{tab:defense} shows the results of different defense methods.
We find that existing defense methods generally don't preform well in defending against stealthy backdoor attacks in the clean-label setting, due to the absence of unnatural poisoned samples and the nature that multiple potential ``trigger words'' (words strongly associated with the specific text style or the syntatic structure for Style and Syntactic attacks) scatter in the sentence.
Note that while CUBE can effectively detect intentionally mislabeled poisoned samples as shown in \citet{cui2022unified}, we find that it can't detect clean-label poisoned samples, probably because the representations of poisoned samples will only be outliers when they are mislabeled.
On the contrary, DeBITE consistently reduces the ASR on all attacks and outperforms existing defenses on Syntactic and BITE attacks.
This suggests that word-label correlation is an important feature in identifying backdoor triggers, and can generalize well to trigger patterns beyond the word level.
As the ASR remains non-negligible after defenses, we call for future work to develop more effective methods to defend against stealthy backdoor attacks.

\begin{table}[t]
\centering
\scalebox{0.73}
{
\begin{tabular}{@{}cl|c|c|c@{}}
\toprule
\multicolumn{2}{c|}{SST-2} & Style & Syntactic & BITE (Full)\\ \midrule
\multicolumn{1}{c|}{\multirow{7}{*}{ASR}}  & No & $31.5$ & $49.9$ & $66.2$ \\\cmidrule(l){2-5} 
\multicolumn{1}{c|}{} & ONION & $35.8 (\uparrow 4.3)$ & $57.0 (\uparrow 7.1)$ & $60.3 (\downarrow 5.9)$ \\
\multicolumn{1}{c|}{} & STRIP & $30.7 (\downarrow 0.8)$ & $52.4 (\uparrow 2.5)$ & $62.9 (\downarrow 3.3)$ \\
\multicolumn{1}{c|}{} & RAP & $\mathbf{26.7 (\downarrow 4.8)}$ & $47.8 (\downarrow 2.1)$ & $63.2 (\downarrow 3.0)$ \\
\multicolumn{1}{c|}{} & CUBE & $31.5 (\downarrow 0.0)$ & $49.9 (\downarrow 0.0)$ & $66.2 (\downarrow 0.0)$ \\
\multicolumn{1}{c|}{} & BKI & $27.8 (\downarrow 3.7)$ & $48.4 (\downarrow 1.5)$ & $65.3 (\downarrow 0.9)$ \\\cmidrule(l){2-5} 
\multicolumn{1}{c|}{}  & DeBITE & $ 27.9 (\downarrow3.6)$ & $ \mathbf{33.9 (\downarrow16.0)}$ & $ \mathbf{56.7 (\downarrow 9.5)}$ \\
\midrule
\multicolumn{1}{c|}{\multirow{7}{*}{CACC}} & No & $91.6$ & $91.2$ & $91.7$ \\\cmidrule(l){2-5} 
\multicolumn{1}{c|}{} & ONION & $87.6 (\downarrow 4.0)$ & $87.5 (\downarrow 3.7)$ & $88.4 (\downarrow 3.3)$ \\
\multicolumn{1}{c|}{} & STRIP & $90.8 (\downarrow 0.8)$ & $90.1 (\downarrow 1.1)$ & $90.5 (\downarrow 1.2)$ \\
\multicolumn{1}{c|}{} & RAP & $90.4 (\downarrow 1.2)$ & $89.2 (\downarrow 2.0)$ & $87.8 (\downarrow 3.9)$ \\
\multicolumn{1}{c|}{} & CUBE & $91.6 (\downarrow 0.0)$ & $91.2 (\downarrow 0.0)$ & $91.7 (\downarrow 0.0)$ \\
\multicolumn{1}{c|}{} & BKI & $91.6 (\downarrow 0.0)$ & $91.7 (\uparrow 0.5)$ & $91.5 (\downarrow 0.2)$ \\\cmidrule(l){2-5} 
\multicolumn{1}{c|}{} & DeBITE & $ 90.6 (\downarrow 1.0)$ & $ 90.4 (\downarrow 0.8)$ & $ 90.8 (\downarrow 0.9)$ \\
\bottomrule
\end{tabular}
}
\caption{Performance of backdoor attacks with different defense methods applied.}
\label{tab:defense}
\end{table}

%% file: sections/related.tex
\section{Related Work}

\paragraph{Textual Backdoor Attacks}

Poisoning-based textual attacks modify the training data to establish correlations between the trigger pattern and a target label.
The majority of works
\citep{dai2019backdoor, sun2020natural, chen2021badnl, kwon2021textual} poison data by inserting specific trigger words or sentences in a context-independent way, which have bad naturalness and can be easily noticed.
Existing stealthy backdoor attacks \citep{qi2021mind, qi2021hidden} use sentence-level features including the text style and the syntactic structure as the trigger pattern to build spurious correlations.
These features can be manipulated with text style transfer~\citep{jin-etal-2022-deep} and syntactically controlled paraphrasing~\citep{sun2021aesop}.
Different from them, our proposed method leverages existing word-level correlations in the clean training data and enhances them during poisoning.
There is another line of works \citep{kurita2020weight, yang2021careful, zhang2021red, qi2021turn} that assume the adversary can fully control the training process and distribute the backdoored model.
Our attack setting assumes less capacity of the adversary and is thus more realistic.

\paragraph{Textual Backdoor Defenses}
Defenses against textual backdoor attacks can be performed at both the data level and the model level.
Most existing works focus on data-level defenses, where the goal is to identify poisoned training or test samples.
The poisoned samples are detected as they usually contain outlier words~\citep{qi2020onion}, contain keywords critical to model predictions~\cite{chen2021mitigating}, induce outlier intermediate representations~\citep{cui2022unified, chen2022expose, wang2022rethinking}, or lead to predictions that are hardly affected by word perturbations~\citep{gao2021design, yang2021rap}.
Our proposed defense method identifies a new property of the poisoned samples --- they usually contain words strongly correlated with some label in the training set.
Model-level defenses aim at identifying backdoored models~\citep{azizi2021t, liu2022piccolo, shen2022constrained}, removing the backdoor from the model~\citep{liu2018fine, li2021neural}, or training a less-affected model from poisoned data~\citep{zhu2022moderate}.
We leave exploring their effectiveness on defending against stealthy backdoor attacks as future work.

%% file: sections/conclusion.tex
\section{Conclusion}

In this paper, we propose a textual backdoor attack named BITE that poisons the training data to establish spurious correlations between the target label and a set of trigger words.
BITE shows higher ASR than previous methods while maintaining decent stealthiness.
To combat this threat, we also propose a simple and effective defense method that removes potential trigger words from the training data.
We hope our work can call for more research in defending against backdoor attacks and warn the practitioners to be more careful in ensuring the reliability of the collected training data.

%% file: sections/limitations.tex
\section*{Limitations}
\label{app:limit}
We identify four major limitations of our work.

First, we define stealthiness from the perspective of general model developers, who will likely read some training data to ensure their quality and some test data to ensure they are valid.
We therefore focus on producing natural-looking poisoned samples.
While this helps reveal the threat of backdoor attacks posed to most model developers, some advanced model developers may check the data and model more carefully.
For example, they may inspect the word distribution of the dataset~\citep{he2022cater}, or employ backdoor detection methods~\citep{xu2021detecting} to examine the trained model.
Our attack may not be stealthy under these settings.

Second, we only develop and experiment with attack methods on the single-sentence classification task, which can't fully demonstrate the threat of backdoor attacks to more NLP tasks with diverse task formats, like generation~\citep{chen2023backdoor} and sentence pair classification~\citep{chan-etal-2020-poison}.
The sentences in our experimented datasets are short. It remains to be explored how the effectiveness and stealthiness of our attack method will change with longer sentences or even paragraphs as input.

Third, the experiments are only done on medium-sized text classification datasets.
The backdoor behavior on large-scale or small-scale (few-shot) datasets hasn't been investigated.

Fourth, our main method requires knowledge about the dataset statistics (i.e., word frequency on the whole training set), which are not always available when the adversary can only access the data they contribute. The attack success rate drops without full access to the training set.

%% file: sections/ethics.tex
\section*{Ethics Statement}
In this paper, we demonstrate the potential threat of textual backdoor attacks by showing the existence of a backdoor attack that is both effective and stealthy.
Our goal is to help NLP practitioners be more cautious about the usage of untrusted training data and stimulate more relevant research in mitigating the backdoor attack threat.

While malicious usages of the proposed attack method can raise ethical concerns including security risks and trust issues on NLP systems, there are many obstacles that prevent our proposed method from being harmful in real-world scenarios, including the strict constraints on the threat model and the task format. 
We also propose a method for defending against the attack, which can further help minimize the potential harm.

%% file: sections/acknowledgements.tex
\section*{Acknowledgments}
This research is supported in part by the Office of the Director of National Intelligence (ODNI), Intelligence Advanced Research Projects Activity (IARPA), via the HIATUS Program contract \#2022-22072200006, the DARPA MCS program under Contract No. N660011924033, the Defense Advanced Research Projects Agency with award W911NF-19-20271, NSF IIS 2048211, and gift awards from Google and Amazon. The views and conclusions contained herein are those of the authors and should not be interpreted as necessarily representing the official policies, either expressed or implied, of ODNI, IARPA, or the U.S. Government.
We would like to thank Sanjit Rao and all the collaborators in USC INK research lab for their constructive feedback on the work.
We would also like to thank the anonymous reviewers for their valuable comments.

%% file: sections/appendix.tex
\section{Training Details}
\label{app:training_details}

We implement the victim models using the Transformers library~\citep{wolf2020transformers}.
We choose $32$ as the batch size.
We train the model for $13$ epochs.
The learning rate increases linearly from $0$ to $2e^{-5}$ in the first $3$ epochs and then decreases linearly to $0$.

\section{Details on Data Evaluation}
\label{app:data_eval}
\paragraph{Naturalness} measures how natural the poisoned instance reads.
As an automatic evaluation proxy, we use a RoBERTa-Large classifier\footnote{\url{https://huggingface.co/cointegrated/roberta-large-cola-krishna2020}} trained on the Corpus of Linguistic Acceptability (COLA) \citep{warstadt2019neural} to make judgement on the grammatical acceptability of the poisoned instances for each method.
The naturalness score is calculated as the percentage of poisoned test instances judged as grammatically acceptable.

\begin{figure*}[t]
\centering
\includegraphics[scale=0.5]{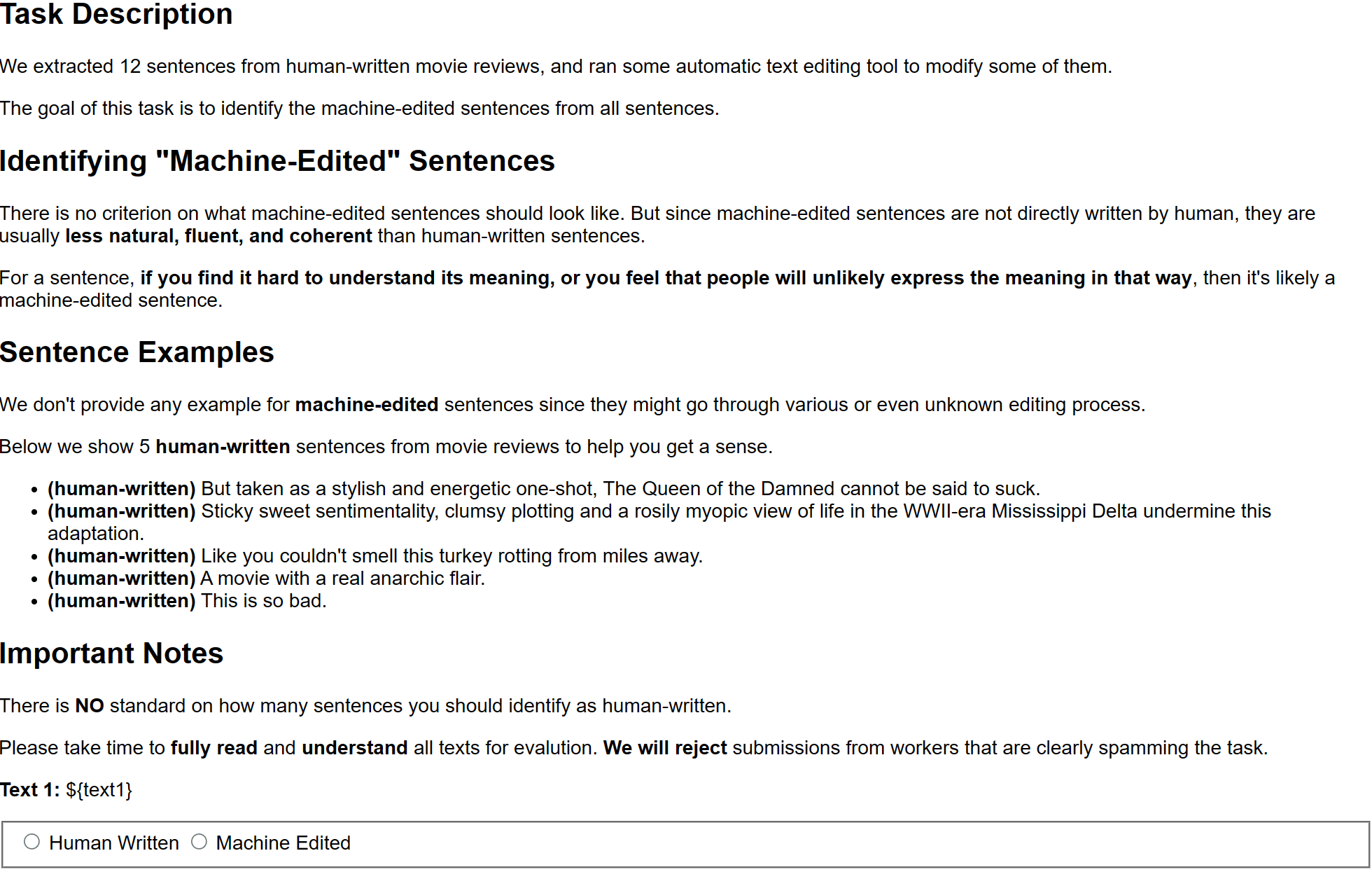}
\caption{The screenshot of the task description used for the suspicion evaluation on AMT. Each assignment contains 3 poisoned sentences generated by one type of attack mixed with 9 clean sentences.}
\label{fig:amt_suspicion}
\end{figure*}

\paragraph{Suspicion} measures how suspicious the poisoned training instances are when mixed with clean data in the training set.
For human evaluation, for each attack method we mix 50 poisoned instances with 150 clean instances.
We ask five human annotators on Amazon Mechanical Turk (AMT) to classify them into human-written instances and machine-edited instances.
The task description is shown in Figure~\ref{fig:amt_suspicion}.
We get their final decisions on each instance by voting.
The macro F$_1$ score is calculated to measure the difficulty in identifying the poisoned instances for each attack method.
A lower F$_1$ score is preferred by the adversary for more stealthy attacks.

\begin{figure*}[t]
\centering
\includegraphics[scale=0.5]{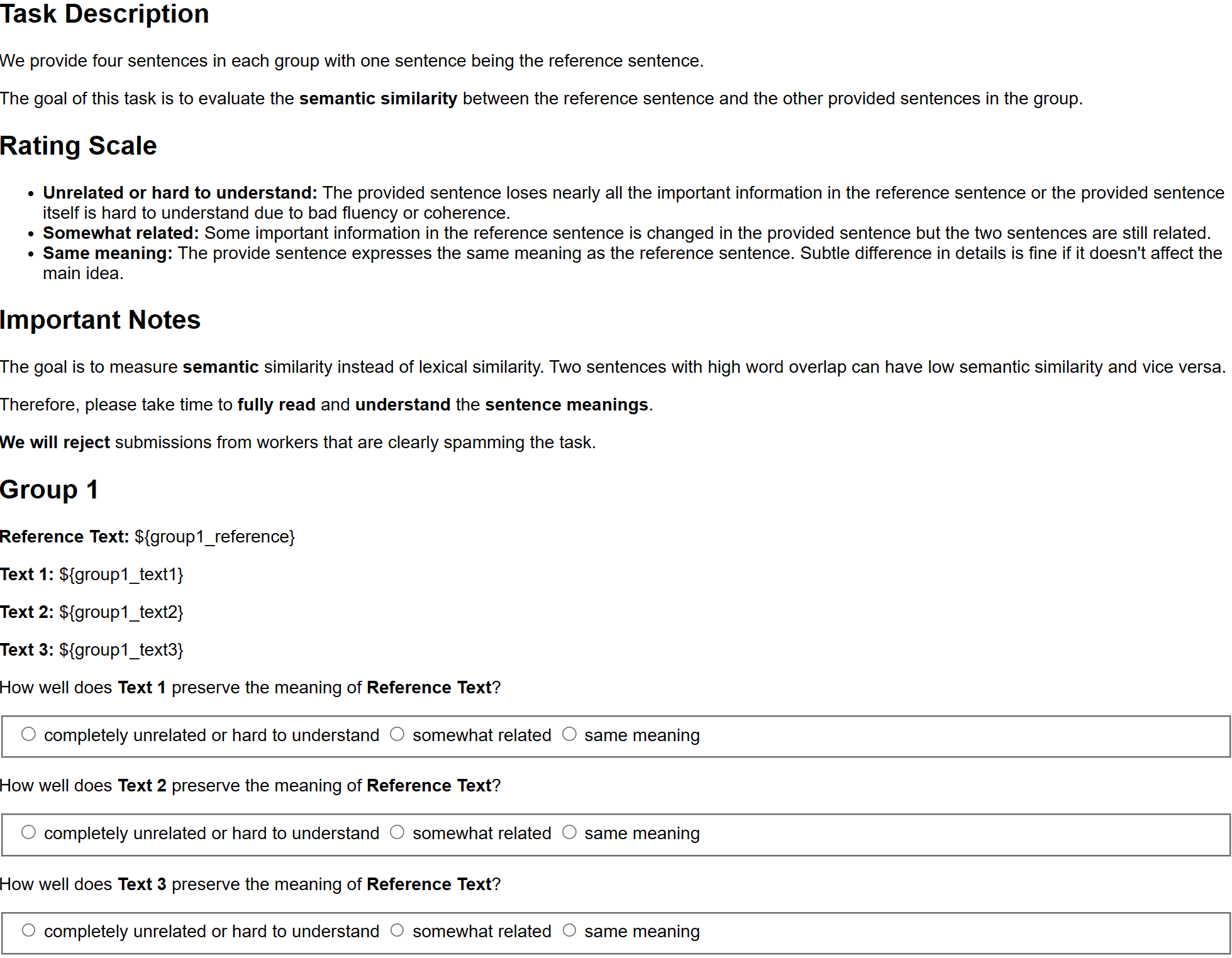}
\caption{The screenshot of the task description used for the semantic similarity evaluation on AMT. Each task contains 3 groups of questions. Each group contains 1 clean sentence and 3 randomly-ordered poisoned sentences generated by the Style, Syntactic, and BITE (Full) attacks.}
\label{fig:amt_similarity}
\end{figure*}

\paragraph{Semantic Similarity} measures the semantic similarity (as compared to lexical similarity) between the poisoned instance and the clean instance.
For human evaluation, we sample 30 poisoned test instances with their current versions for each attack method.
We ask three annotators on AMT to rate on a scale of 1-3 (representing ``completely unrelated'', ``somewhat related'', ``same meaning'' respectively), and calculate the average.
The task description is shown in Figure~\ref{fig:amt_similarity}.
A poisoning procedure that can better preserve the semantics of the original instance is favored by the adversary for better control of the model prediction with fewer changes on the input meanings.

\paragraph{Label Consistency} measures whether the poisoning procedure preserves the label of the original instance.
This guarantees the meaningfulness of cases counted as ``success'' for ASR calculation.
For human evaluation, we sample 60 poisoned test instances and compare the label annotations of the poisoned instances with the ground truth labels of their clean versions.
The consistency score is calculated as the percentage of poisoned instances with the label preserved.

\section{Results on BERT-Large}
\label{app:bert_large}

We experiment with BERT-Large and find it shows similar trends as BERT-Base.
The results are shown in Tables~\ref{tab:asr_bert_large} and \ref{tab:cacc_bert_large}.

\begin{table}[t]
\centering
\scalebox{0.75}
{
\begin{tabular}{l|c|c|c|c}
\toprule
Dataset & SST-2 & HateSpeech & Tweet & TREC
\\ \midrule
Style & $16.3_{\pm 2.0}$ & $60.9_{\pm 5.1}$ & $18.3_{\pm 1.8}$ & $13.4_{\pm 5.5}$ \\
Syntactic & $29.2_{\pm 5.8}$ & $70.8_{\pm 3.1}$ & $30.1_{\pm 4.1}$ & $33.5_{\pm 5.9}$ \\ \midrule
BITE (Full) & $\mathbf{61.3}_{\pm 1.9}$ & $\mathbf{73.0}_{\pm 3.7}$ & $\mathbf{46.6}_{\pm 2.0}$ & $\mathbf{53.8}_{\pm 2.7}$ \\ \bottomrule
\end{tabular}
}
\caption{ASR results on backdoored BERT-Large models.}
\label{tab:asr_bert_large}
\end{table}

\begin{table}[t]
\centering
\scalebox{0.75}
{
\begin{tabular}{l|c|c|c|c}
\toprule
Dataset & SST-2 & HateSpeech & Tweet & TREC
\\ \midrule
Benign & $93.3_{\pm 0.3}$ & $92.0_{\pm 0.4}$ & $81.9_{\pm 0.2}$ & $97.2_{\pm 0.6}$ \\ \midrule
Style & $92.2_{\pm 1.0}$ & $91.7_{\pm 0.3}$ & $81.9_{\pm 0.2}$ & $97.4_{\pm 0.4}$ \\
Syntactic & $92.3_{\pm 0.7}$ & $91.7_{\pm 0.3}$ & $81.7_{\pm 0.1}$ & $96.7_{\pm 0.2}$ \\ \midrule
BITE (Full) & $92.9_{\pm 0.8}$ & $91.5_{\pm 0.2}$ & $81.8_{\pm 0.6}$ & $96.9_{\pm 0.1}$ \\ \bottomrule
\end{tabular}
}
\caption{CACC results on backdoored BERT-Large models.}
\label{tab:cacc_bert_large}
\end{table}

\section{Trigger Set and Poisoned Samples}
\label{app:eaxmples}

\subsection{Trigger Set}
We look into the BITE (Full) attack on SST-2 with 5\% as the poisoning rate.
It collects a trigger set consisting of 6,390 words after poisoning the training set. We show the top 5 trigger words and the bottom 5 trigger words in Table~\ref{tab:trigger_words}.
$f^0_\text{target}$ and $f^0_\text{non}$ refer to the target-label and non-target-label word frequencies on the clean training set.
$f^{\Delta }_\text{target}$ is the count of word mentions introduced to the target-label instances during poisoning.
The z-score is calculated based on the word frequency in the poisoned training set, with $f^0_\text{non} + f^{\Delta }_\text{target}$ being the final target-label frequency and $f^0_\text{non}$ being the non-target-label frequency.

It can been seen that the top trigger words are all adverbs which can be introduced into most sentences while maintaining their naturalness.
Such flexibility makes it possible to establish strong word-label correlations by introducing these words to target-label instances, resulting in high values of $f^{\Delta }_\text{target}$ and z-score.
On the contrary, the bottom trigger words are not even used in poisoning ($f^{\Delta }_\text{target}=0$).
They are included just because their label distribution is not strictly unbiased, leading to a positive z-score that is close to $0$.
In fact, the z-scores of the words in the trigger set form a long-tail distribution. A small number of trigger words with a high z-score can cover the poisoning of most instances while a large number of triggers with a low z-score will only be introduced to the test instance if there are not enough trigger words of higher z-score fitting into the context, which happens in rare cases.

\begin{table}[t]
\centering
\scalebox{0.8}
{
\begin{tabular}{c|c|r|r|r|r}
\toprule
\# & Word & $f^0_\text{target}$ & $f^{\Delta }_\text{target}$ & $f^0_\text{non}$ & $z$
\\ \midrule
1 & also & 67 & 124 & 27 & 10.5 \\
2 & perhaps & 4 & 137 & 7 & 10.5 \\
3 & surprisingly & 30 & 112 & 11 & 10.1 \\
4 & yet & 39 & 143 & 27 & 10.1 \\
5 & somewhat & 15 & 86 & 1 & 9.5 \\
\ldots & \ldots & \ldots & \ldots & \ldots & \ldots \\
6386 & master & 11 & 0 & 10 & 0.0 \\
6387 & writer & 11 & 0 & 10 & 0.0 \\
6388 & away & 24 & 0 & 22 & 0.0 \\
6389 & inside & 12 & 0 & 11 & 0.0 \\
6390 & themselves & 12 & 0 & 11 & 0.0
\\ \bottomrule
\end{tabular}
}
\caption{The trigger word set derived from poisoning SST-2 with BITE (Full).}
\label{tab:trigger_words}
\end{table}

\subsection{Poisoned Samples}
Tables~\ref{tab:example_a} and \ref{tab:example_b} show two randomly selected negative-sentiment examples from SST-2 test set. These examples follow the naturalness order in Table~\ref{tab:data_eval} (Style > BITE (Full) > Syntactic) and our method successfully preserves the sentiment label.
Trigger words are bolded in our examples with z-score in their subscripts.
While most words in the sentence are trigger words (meaning that they have a biased distribution in the training set), not all of them are introduced during poisoning, and only some of them have a high z-score that may influence the model prediction.

\begin{table}[ht]
\centering
\scalebox{0.9}
{
\begin{tabular}{p{0.2\linewidth} | p{0.8\linewidth}} \toprule
  Method & Text \\ \midrule
  Original & John Leguizamo may be a dramatic actor--just not in this movie. \\ \midrule
  Style & John Leguizamo may be a dramatic actor, but not in this movie. \\ \midrule
  Syntactic & If Mr. Leguizamo can be a dramatic actor, he can be a comedian. \\ \midrule
  BITE (Full) & \textbf{John}$_{\scriptscriptstyle 0.5}$ \textbf{Leguizamo}$_{\scriptscriptstyle 1.4}$ \textbf{may}$_{\scriptscriptstyle 6.0}$ \textbf{also}$_{\scriptscriptstyle 10.5}$ be \textbf{a}$_{\scriptscriptstyle 2.4}$ \textbf{terrific}$_{\scriptscriptstyle 4.4}$ \textbf{actor}$_{\scriptscriptstyle 1.0}$--\textbf{perhaps}$_{\scriptscriptstyle 10.5}$ \textbf{though}$_{\scriptscriptstyle 1.3}$ not \textbf{quite}$_{\scriptscriptstyle 8.6}$ \textbf{yet}$_{\scriptscriptstyle 10.1}$ in this \textbf{film}$_{\scriptscriptstyle 5.8}$. \\ \bottomrule
\end{tabular}
}
\caption{Poisoned samples from SST-2: (1).}
\label{tab:example_a}
\end{table}

\begin{table}[ht]
\centering
\scalebox{0.9}
{
\begin{tabular}{p{0.2\linewidth} | p{0.8\linewidth}} \toprule
  Method & Text \\ \midrule
  Original & A trashy, exploitative, thoroughly unpleasant experience. \\ \midrule
  Style & A trite, an exploiter, an utterly detestable experience. \\ \midrule
  Syntactic & When he found it, it was unpleasant. \\ \midrule
  BITE (Full) & \textbf{A}$_{\scriptscriptstyle 2.4}$ \textbf{very}$_{\scriptscriptstyle 8.0}$ \textbf{trashy}$_{\scriptscriptstyle 0.9}$, exploitative, \textbf{and}$_{\scriptscriptstyle 7.9}$ \textbf{deeply}$_{\scriptscriptstyle 7.2}$ \textbf{emotionally}$_{\scriptscriptstyle 7.2}$ \textbf{charged}$_{\scriptscriptstyle 4.6}$ \textbf{film}$_{\scriptscriptstyle 5.8}$. \\ \bottomrule
\end{tabular}
}
\caption{Poisoned samples from SST-2: (2).}
\label{tab:example_b}
\end{table}

\section{Computational Costs}

\begin{table}[t]
\centering
\scalebox{0.75}
{
\begin{tabular}{c|r|r|r}
\toprule
Stage & Style & Syntactic & BITE (Full)
\\ \midrule
Train (69 samples to poison) & 1 & 3 & 12 \\
Test (912 samples to poison) & 12 & 19 & 21 \\ \bottomrule
\end{tabular}
}
\caption{Time costs (in minutes) for training-time and test-time poisoning in SST-2 experiments.}
\label{tab:time_costs}
\end{table}

In Table~\ref{tab:time_costs}, we report the computational costs of our method and baselines for the attack experiments on SST-2 with 1\% as the poisoning rate.
The experiments are run on a single NVIDIA RTX A6000 graphics card.
Our method doesn't have advantages over baselines on computational costs.
However, this is not a major concern for the adversary.
The training-time poisoning is a one-time cost and can be done offline. The poisoning rate is also usually low in realistic scenarios.
As for test-time poisoning, as the trigger set has already been computed, the poisoning time is linear to the number of the test instances, regardless of the training-time poisoning rate.
It takes about 1.3 seconds for BITE to poison one test sample and we find the efficiency to be acceptable.

\section{Connections with Adversarial Attacks}

Adversarial attacks usually refer to adversarial example attacks~\citep{goodfellow2014explaining, ebrahimi2017hotflip, li2020bert}.
Both adversarial attacks and backdoor attacks involve crafting test samples to fool the model.
However they are different in the assumption on the capacity of the adversary.
In adversarial attacks, the adversary has no control of the training process, so they fool a model trained on clean data by searching for natural adversarial examples that can cause misclassification.
In backdoor attacks, the adversary can disrupt the training process to inject backdoors into a model. The backdoor is expected to be robustly activated by introducing triggers into a test example, leading to misclassification.
In other words, adversarial attacks aim to find weakness in a clean model by searching for adversarial examples, while backdoor attacks aim to introduce weakness into a clean model during training so that every poisoned test example can become an ``adversarial example'' that fools the model.
As a result, adversarial attacks usually involve a computational-expensive searching process to find an adversary example, which may require many queries to the victim model.
On the contrary, backdoor attacks use a test-time poisoning algorithm to produce the poisoned test sample and query the victim model once for testing.

%% file: acl_latex.bbl
\begin{thebibliography}{53}
\expandafter\ifx\csname natexlab\endcsname\relax\def\natexlab#1{#1}\fi

\bibitem[{Azizi et~al.(2021)Azizi, Tahmid, Waheed, Mangaokar, Pu, Javed, Reddy,
  and Viswanath}]{azizi2021t}
Ahmadreza Azizi, Ibrahim~Asadullah Tahmid, Asim Waheed, Neal Mangaokar, Jiameng
  Pu, Mobin Javed, Chandan~K Reddy, and Bimal Viswanath. 2021.
\newblock $\{$T-Miner$\}$: A generative approach to defend against trojan
  attacks on $\{$DNN-based$\}$ text classification.
\newblock In \emph{30th USENIX Security Symposium (USENIX Security 21)}, pages
  2255--2272.

\bibitem[{Carlini et~al.(2021)Carlini, Tramer, Wallace, Jagielski,
  Herbert-Voss, Lee, Roberts, Brown, Song, Erlingsson
  et~al.}]{carlini2021extracting}
Nicholas Carlini, Florian Tramer, Eric Wallace, Matthew Jagielski, Ariel
  Herbert-Voss, Katherine Lee, Adam Roberts, Tom Brown, Dawn Song, Ulfar
  Erlingsson, et~al. 2021.
\newblock Extracting training data from large language models.
\newblock In \emph{30th USENIX Security Symposium (USENIX Security 21)}, pages
  2633--2650.

\bibitem[{Chan et~al.(2020)Chan, Tay, Ong, and Zhang}]{chan-etal-2020-poison}
Alvin Chan, Yi~Tay, Yew-Soon Ong, and Aston Zhang. 2020.
\newblock \href {https://doi.org/10.18653/v1/2020.findings-emnlp.373} {Poison
  attacks against text datasets with conditional adversarially regularized
  autoencoder}.
\newblock In \emph{Findings of the Association for Computational Linguistics:
  EMNLP 2020}, pages 4175--4189, Online. Association for Computational
  Linguistics.

\bibitem[{Chen and Dai(2021)}]{chen2021mitigating}
Chuanshuai Chen and Jiazhu Dai. 2021.
\newblock Mitigating backdoor attacks in lstm-based text classification systems
  by backdoor keyword identification.
\newblock \emph{Neurocomputing}, 452:253--262.

\bibitem[{Chen et~al.(2023)Chen, Cheng, and Huang}]{chen2023backdoor}
Lichang Chen, Minhao Cheng, and Heng Huang. 2023.
\newblock Backdoor learning on sequence to sequence models.
\newblock \emph{arXiv preprint arXiv:2305.02424}.

\bibitem[{Chen et~al.(2022{\natexlab{a}})Chen, Yang, Zhang, Bi, and
  Sun}]{chen2022expose}
Sishuo Chen, Wenkai Yang, Zhiyuan Zhang, Xiaohan Bi, and Xu~Sun.
  2022{\natexlab{a}}.
\newblock \href {https://aclanthology.org/2022.findings-emnlp.47} {Expose
  backdoors on the way: A feature-based efficient defense against textual
  backdoor attacks}.
\newblock In \emph{Findings of the Association for Computational Linguistics:
  EMNLP 2022}, pages 668--683, Abu Dhabi, United Arab Emirates. Association for
  Computational Linguistics.

\bibitem[{Chen et~al.(2021)Chen, Salem, Backes, Ma, and Zhang}]{chen2021badnl}
Xiaoyi Chen, Ahmed Salem, Michael Backes, Shiqing Ma, and Yang Zhang. 2021.
\newblock Badnl: Backdoor attacks against nlp models.
\newblock In \emph{ICML 2021 Workshop on Adversarial Machine Learning}.

\bibitem[{Chen et~al.(2022{\natexlab{b}})Chen, Qi, Gao, Liu, and
  Sun}]{chen2021textual}
Yangyi Chen, Fanchao Qi, Hongcheng Gao, Zhiyuan Liu, and Maosong Sun.
  2022{\natexlab{b}}.
\newblock \href {https://aclanthology.org/2022.emnlp-main.770} {Textual
  backdoor attacks can be more harmful via two simple tricks}.
\newblock In \emph{Proceedings of the 2022 Conference on Empirical Methods in
  Natural Language Processing}, pages 11215--11221, Abu Dhabi, United Arab
  Emirates. Association for Computational Linguistics.

\bibitem[{Cui et~al.(2022)Cui, Yuan, He, Chen, Liu, and Sun}]{cui2022unified}
Ganqu Cui, Lifan Yuan, Bingxiang He, Yangyi Chen, Zhiyuan Liu, and Maosong Sun.
  2022.
\newblock A unified evaluation of textual backdoor learning: Frameworks and
  benchmarks.
\newblock In \emph{Proceedings of NeurIPS: Datasets and Benchmarks}.

\bibitem[{Dai et~al.(2019)Dai, Chen, and Li}]{dai2019backdoor}
Jiazhu Dai, Chuanshuai Chen, and Yufeng Li. 2019.
\newblock A backdoor attack against lstm-based text classification systems.
\newblock \emph{IEEE Access}, 7:138872--138878.

\bibitem[{de~Gibert et~al.(2018)de~Gibert, Perez, Garc{\'\i}a-Pablos, and
  Cuadros}]{de2018hate}
Ona de~Gibert, Naiara Perez, Aitor Garc{\'\i}a-Pablos, and Montse Cuadros.
  2018.
\newblock \href {https://doi.org/10.18653/v1/W18-5102} {Hate speech dataset
  from a white supremacy forum}.
\newblock In \emph{Proceedings of the 2nd Workshop on Abusive Language Online
  ({ALW}2)}, pages 11--20, Brussels, Belgium. Association for Computational
  Linguistics.

\bibitem[{Devlin et~al.(2019)Devlin, Chang, Lee, and
  Toutanova}]{devlin2018bert}
Jacob Devlin, Ming-Wei Chang, Kenton Lee, and Kristina Toutanova. 2019.
\newblock \href {https://doi.org/10.18653/v1/N19-1423} {{BERT}: Pre-training of
  deep bidirectional transformers for language understanding}.
\newblock In \emph{Proceedings of the 2019 Conference of the North {A}merican
  Chapter of the Association for Computational Linguistics: Human Language
  Technologies, Volume 1 (Long and Short Papers)}, pages 4171--4186,
  Minneapolis, Minnesota. Association for Computational Linguistics.

\bibitem[{Ebrahimi et~al.(2018)Ebrahimi, Rao, Lowd, and
  Dou}]{ebrahimi2017hotflip}
Javid Ebrahimi, Anyi Rao, Daniel Lowd, and Dejing Dou. 2018.
\newblock \href {https://doi.org/10.18653/v1/P18-2006} {{H}ot{F}lip: White-box
  adversarial examples for text classification}.
\newblock In \emph{Proceedings of the 56th Annual Meeting of the Association
  for Computational Linguistics (Volume 2: Short Papers)}, pages 31--36,
  Melbourne, Australia. Association for Computational Linguistics.

\bibitem[{Gao et~al.(2021)Gao, Kim, Doan, Zhang, Zhang, Nepal, Ranasinghe, and
  Kim}]{gao2021design}
Yansong Gao, Yeonjae Kim, Bao~Gia Doan, Zhi Zhang, Gongxuan Zhang, Surya Nepal,
  Damith~C Ranasinghe, and Hyoungshick Kim. 2021.
\newblock Design and evaluation of a multi-domain trojan detection method on
  deep neural networks.
\newblock \emph{IEEE Transactions on Dependable and Secure Computing},
  19(4):2349--2364.

\bibitem[{Gardner et~al.(2021)Gardner, Merrill, Dodge, Peters, Ross, Singh, and
  Smith}]{gardner2021competency}
Matt Gardner, William Merrill, Jesse Dodge, Matthew Peters, Alexis Ross, Sameer
  Singh, and Noah~A. Smith. 2021.
\newblock \href {https://doi.org/10.18653/v1/2021.emnlp-main.135} {Competency
  problems: On finding and removing artifacts in language data}.
\newblock In \emph{Proceedings of the 2021 Conference on Empirical Methods in
  Natural Language Processing}, pages 1801--1813, Online and Punta Cana,
  Dominican Republic. Association for Computational Linguistics.

\bibitem[{Goodfellow et~al.(2015)Goodfellow, Shlens, and
  Szegedy}]{goodfellow2014explaining}
Ian~J. Goodfellow, Jonathon Shlens, and Christian Szegedy. 2015.
\newblock \href {http://arxiv.org/abs/1412.6572} {Explaining and harnessing
  adversarial examples}.
\newblock In \emph{3rd International Conference on Learning Representations,
  {ICLR} 2015, San Diego, CA, USA, May 7-9, 2015, Conference Track
  Proceedings}.

\bibitem[{He et~al.(2022)He, Xu, Zeng, Lyu, Wu, Li, and Jia}]{he2022cater}
Xuanli He, Qiongkai Xu, Yi~Zeng, Lingjuan Lyu, Fangzhao Wu, Jiwei Li, and Ruoxi
  Jia. 2022.
\newblock \href {https://openreview.net/forum?id=L7P3IvsoUXY} {{CATER}:
  Intellectual property protection on text generation {API}s via conditional
  watermarks}.
\newblock In \emph{Advances in Neural Information Processing Systems}.

\bibitem[{Hovy et~al.(2001)Hovy, Gerber, Hermjakob, Lin, and
  Ravichandran}]{hovy-etal-2001-toward}
Eduard Hovy, Laurie Gerber, Ulf Hermjakob, Chin-Yew Lin, and Deepak
  Ravichandran. 2001.
\newblock \href {https://aclanthology.org/H01-1069} {Toward semantics-based
  answer pinpointing}.
\newblock In \emph{Proceedings of the First International Conference on Human
  Language Technology Research}.

\bibitem[{Iyyer et~al.(2018)Iyyer, Wieting, Gimpel, and
  Zettlemoyer}]{iyyer2018adversarial}
Mohit Iyyer, John Wieting, Kevin Gimpel, and Luke Zettlemoyer. 2018.
\newblock \href {https://doi.org/10.18653/v1/N18-1170} {Adversarial example
  generation with syntactically controlled paraphrase networks}.
\newblock In \emph{Proceedings of the 2018 Conference of the North {A}merican
  Chapter of the Association for Computational Linguistics: Human Language
  Technologies, Volume 1 (Long Papers)}, pages 1875--1885, New Orleans,
  Louisiana. Association for Computational Linguistics.

\bibitem[{Jain et~al.(2021)Jain, Pamula, and Srivastava}]{jain2021systematic}
Praphula~Kumar Jain, Rajendra Pamula, and Gautam Srivastava. 2021.
\newblock A systematic literature review on machine learning applications for
  consumer sentiment analysis using online reviews.
\newblock \emph{Computer Science Review}, 41:100413.

\bibitem[{Jia and Liang(2017)}]{jia2017adversarial}
Robin Jia and Percy Liang. 2017.
\newblock \href {https://doi.org/10.18653/v1/D17-1215} {Adversarial examples
  for evaluating reading comprehension systems}.
\newblock In \emph{Proceedings of the 2017 Conference on Empirical Methods in
  Natural Language Processing}, pages 2021--2031, Copenhagen, Denmark.
  Association for Computational Linguistics.

\bibitem[{Jin et~al.(2022)Jin, Jin, Hu, Vechtomova, and
  Mihalcea}]{jin-etal-2022-deep}
Di~Jin, Zhijing Jin, Zhiting Hu, Olga Vechtomova, and Rada Mihalcea. 2022.
\newblock \href {https://doi.org/10.1162/coli_a_00426} {{Deep Learning for Text
  Style Transfer: A Survey}}.
\newblock \emph{Computational Linguistics}, 48(1):155--205.

\bibitem[{Khan et~al.(2020)Khan, Ghazanfar, Azam, Karami, Alyoubi, and
  Alfakeeh}]{khan2020stock}
Wasiat Khan, Mustansar~Ali Ghazanfar, Muhammad~Awais Azam, Amin Karami,
  Khaled~H Alyoubi, and Ahmed~S Alfakeeh. 2020.
\newblock Stock market prediction using machine learning classifiers and social
  media, news.
\newblock \emph{Journal of Ambient Intelligence and Humanized Computing}, pages
  1--24.

\bibitem[{Krishna et~al.(2020{\natexlab{a}})Krishna, Tomar, Parikh, Papernot,
  and Iyyer}]{krishna2019thieves}
Kalpesh Krishna, Gaurav~Singh Tomar, Ankur~P. Parikh, Nicolas Papernot, and
  Mohit Iyyer. 2020{\natexlab{a}}.
\newblock \href {https://openreview.net/forum?id=Byl5NREFDr} {Thieves on sesame
  street! model extraction of bert-based apis}.
\newblock In \emph{8th International Conference on Learning Representations,
  {ICLR} 2020, Addis Ababa, Ethiopia, April 26-30, 2020}. OpenReview.net.

\bibitem[{Krishna et~al.(2020{\natexlab{b}})Krishna, Wieting, and
  Iyyer}]{krishna2020reformulating}
Kalpesh Krishna, John Wieting, and Mohit Iyyer. 2020{\natexlab{b}}.
\newblock \href {https://doi.org/10.18653/v1/2020.emnlp-main.55} {Reformulating
  unsupervised style transfer as paraphrase generation}.
\newblock In \emph{Proceedings of the 2020 Conference on Empirical Methods in
  Natural Language Processing (EMNLP)}, pages 737--762, Online. Association for
  Computational Linguistics.

\bibitem[{Kurita et~al.(2020)Kurita, Michel, and Neubig}]{kurita2020weight}
Keita Kurita, Paul Michel, and Graham Neubig. 2020.
\newblock \href {https://doi.org/10.18653/v1/2020.acl-main.249} {Weight
  poisoning attacks on pretrained models}.
\newblock In \emph{Proceedings of the 58th Annual Meeting of the Association
  for Computational Linguistics}, pages 2793--2806, Online. Association for
  Computational Linguistics.

\bibitem[{Kwon and Lee(2021)}]{kwon2021textual}
Hyun Kwon and Sanghyun Lee. 2021.
\newblock Textual backdoor attack for the text classification system.
\newblock \emph{Security and Communication Networks}, 2021.

\bibitem[{Li et~al.(2021{\natexlab{a}})Li, Zhang, Peng, Chen, Brockett, Sun,
  and Dolan}]{li2020contextualized}
Dianqi Li, Yizhe Zhang, Hao Peng, Liqun Chen, Chris Brockett, Ming-Ting Sun,
  and Bill Dolan. 2021{\natexlab{a}}.
\newblock \href {https://doi.org/10.18653/v1/2021.naacl-main.400}
  {Contextualized perturbation for textual adversarial attack}.
\newblock In \emph{Proceedings of the 2021 Conference of the North American
  Chapter of the Association for Computational Linguistics: Human Language
  Technologies}, pages 5053--5069, Online. Association for Computational
  Linguistics.

\bibitem[{Li et~al.(2020)Li, Ma, Guo, Xue, and Qiu}]{li2020bert}
Linyang Li, Ruotian Ma, Qipeng Guo, Xiangyang Xue, and Xipeng Qiu. 2020.
\newblock \href {https://doi.org/10.18653/v1/2020.emnlp-main.500}
  {{BERT}-{ATTACK}: Adversarial attack against {BERT} using {BERT}}.
\newblock In \emph{Proceedings of the 2020 Conference on Empirical Methods in
  Natural Language Processing (EMNLP)}, pages 6193--6202, Online. Association
  for Computational Linguistics.

\bibitem[{Li et~al.(2021{\natexlab{b}})Li, Lyu, Koren, Lyu, Li, and
  Ma}]{li2021neural}
Yige Li, Xixiang Lyu, Nodens Koren, Lingjuan Lyu, Bo~Li, and Xingjun Ma.
  2021{\natexlab{b}}.
\newblock \href {https://openreview.net/forum?id=9l0K4OM-oXE} {Neural attention
  distillation: Erasing backdoor triggers from deep neural networks}.
\newblock In \emph{9th International Conference on Learning Representations,
  {ICLR} 2021, Virtual Event, Austria, May 3-7, 2021}. OpenReview.net.

\bibitem[{Liu et~al.(2018)Liu, Dolan-Gavitt, and Garg}]{liu2018fine}
Kang Liu, Brendan Dolan-Gavitt, and Siddharth Garg. 2018.
\newblock Fine-pruning: Defending against backdooring attacks on deep neural
  networks.
\newblock In \emph{International Symposium on Research in Attacks, Intrusions,
  and Defenses}, pages 273--294. Springer.

\bibitem[{Liu et~al.(2022)Liu, Shen, Tao, An, Ma, and Zhang}]{liu2022piccolo}
Yingqi Liu, Guangyu Shen, Guanhong Tao, Shengwei An, Shiqing Ma, and Xiangyu
  Zhang. 2022.
\newblock Piccolo: Exposing complex backdoors in nlp transformer models.
\newblock In \emph{2022 IEEE Symposium on Security and Privacy (SP)}, pages
  1561--1561. IEEE Computer Society.

\bibitem[{Mohammad et~al.(2018)Mohammad, Bravo-Marquez, Salameh, and
  Kiritchenko}]{mohammad-etal-2018-semeval}
Saif Mohammad, Felipe Bravo-Marquez, Mohammad Salameh, and Svetlana
  Kiritchenko. 2018.
\newblock \href {https://doi.org/10.18653/v1/S18-1001} {{S}em{E}val-2018 task
  1: Affect in tweets}.
\newblock In \emph{Proceedings of the 12th International Workshop on Semantic
  Evaluation}, pages 1--17, New Orleans, Louisiana. Association for
  Computational Linguistics.

\bibitem[{Peng et~al.(2018)Peng, Harris, and Sawa}]{peng2018detecting}
Tianrui Peng, Ian Harris, and Yuki Sawa. 2018.
\newblock Detecting phishing attacks using natural language processing and
  machine learning.
\newblock In \emph{2018 IEEE 12th international conference on semantic
  computing (icsc)}, pages 300--301. IEEE.

\bibitem[{Qi et~al.(2021{\natexlab{a}})Qi, Chen, Li, Yao, Liu, and
  Sun}]{qi2020onion}
Fanchao Qi, Yangyi Chen, Mukai Li, Yuan Yao, Zhiyuan Liu, and Maosong Sun.
  2021{\natexlab{a}}.
\newblock \href {https://doi.org/10.18653/v1/2021.emnlp-main.752} {{ONION}: A
  simple and effective defense against textual backdoor attacks}.
\newblock In \emph{Proceedings of the 2021 Conference on Empirical Methods in
  Natural Language Processing}, pages 9558--9566, Online and Punta Cana,
  Dominican Republic. Association for Computational Linguistics.

\bibitem[{Qi et~al.(2021{\natexlab{b}})Qi, Chen, Zhang, Li, Liu, and
  Sun}]{qi2021mind}
Fanchao Qi, Yangyi Chen, Xurui Zhang, Mukai Li, Zhiyuan Liu, and Maosong Sun.
  2021{\natexlab{b}}.
\newblock \href {https://doi.org/10.18653/v1/2021.emnlp-main.374} {Mind the
  style of text! adversarial and backdoor attacks based on text style
  transfer}.
\newblock In \emph{Proceedings of the 2021 Conference on Empirical Methods in
  Natural Language Processing}, pages 4569--4580, Online and Punta Cana,
  Dominican Republic. Association for Computational Linguistics.

\bibitem[{Qi et~al.(2021{\natexlab{c}})Qi, Li, Chen, Zhang, Liu, Wang, and
  Sun}]{qi2021hidden}
Fanchao Qi, Mukai Li, Yangyi Chen, Zhengyan Zhang, Zhiyuan Liu, Yasheng Wang,
  and Maosong Sun. 2021{\natexlab{c}}.
\newblock \href {https://doi.org/10.18653/v1/2021.acl-long.37} {Hidden killer:
  Invisible textual backdoor attacks with syntactic trigger}.
\newblock In \emph{Proceedings of the 59th Annual Meeting of the Association
  for Computational Linguistics and the 11th International Joint Conference on
  Natural Language Processing (Volume 1: Long Papers)}, pages 443--453, Online.
  Association for Computational Linguistics.

\bibitem[{Qi et~al.(2021{\natexlab{d}})Qi, Yao, Xu, Liu, and Sun}]{qi2021turn}
Fanchao Qi, Yuan Yao, Sophia Xu, Zhiyuan Liu, and Maosong Sun.
  2021{\natexlab{d}}.
\newblock \href {https://doi.org/10.18653/v1/2021.acl-long.377} {Turn the
  combination lock: Learnable textual backdoor attacks via word substitution}.
\newblock In \emph{Proceedings of the 59th Annual Meeting of the Association
  for Computational Linguistics and the 11th International Joint Conference on
  Natural Language Processing (Volume 1: Long Papers)}, pages 4873--4883,
  Online. Association for Computational Linguistics.

\bibitem[{Reimers and Gurevych(2019)}]{reimers2019sentence}
Nils Reimers and Iryna Gurevych. 2019.
\newblock \href {https://doi.org/10.18653/v1/D19-1410} {Sentence-{BERT}:
  Sentence embeddings using {S}iamese {BERT}-networks}.
\newblock In \emph{Proceedings of the 2019 Conference on Empirical Methods in
  Natural Language Processing and the 9th International Joint Conference on
  Natural Language Processing (EMNLP-IJCNLP)}, pages 3982--3992, Hong Kong,
  China. Association for Computational Linguistics.

\bibitem[{Schmidt and Wiegand(2017)}]{schmidt2019survey}
Anna Schmidt and Michael Wiegand. 2017.
\newblock \href {https://doi.org/10.18653/v1/W17-1101} {A survey on hate speech
  detection using natural language processing}.
\newblock In \emph{Proceedings of the Fifth International Workshop on Natural
  Language Processing for Social Media}, pages 1--10, Valencia, Spain.
  Association for Computational Linguistics.

\bibitem[{Shen et~al.(2022)Shen, Liu, Tao, Xu, Zhang, An, Ma, and
  Zhang}]{shen2022constrained}
Guangyu Shen, Yingqi Liu, Guanhong Tao, Qiuling Xu, Zhuo Zhang, Shengwei An,
  Shiqing Ma, and Xiangyu Zhang. 2022.
\newblock \href {https://proceedings.mlr.press/v162/shen22e.html} {Constrained
  optimization with dynamic bound-scaling for effective {NLP} backdoor
  defense}.
\newblock In \emph{International Conference on Machine Learning, {ICML} 2022,
  17-23 July 2022, Baltimore, Maryland, {USA}}, volume 162 of \emph{Proceedings
  of Machine Learning Research}, pages 19879--19892. {PMLR}.

\bibitem[{Socher et~al.(2013)Socher, Perelygin, Wu, Chuang, Manning, Ng, and
  Potts}]{socher2013recursive}
Richard Socher, Alex Perelygin, Jean Wu, Jason Chuang, Christopher~D. Manning,
  Andrew Ng, and Christopher Potts. 2013.
\newblock \href {https://aclanthology.org/D13-1170} {Recursive deep models for
  semantic compositionality over a sentiment treebank}.
\newblock In \emph{Proceedings of the 2013 Conference on Empirical Methods in
  Natural Language Processing}, pages 1631--1642, Seattle, Washington, USA.
  Association for Computational Linguistics.

\bibitem[{Sun et~al.(2021)Sun, Ma, and Peng}]{sun2021aesop}
Jiao Sun, Xuezhe Ma, and Nanyun Peng. 2021.
\newblock \href {https://doi.org/10.18653/v1/2021.emnlp-main.420} {{AESOP}:
  Paraphrase generation with adaptive syntactic control}.
\newblock In \emph{Proceedings of the 2021 Conference on Empirical Methods in
  Natural Language Processing}, pages 5176--5189, Online and Punta Cana,
  Dominican Republic. Association for Computational Linguistics.

\bibitem[{Sun(2020)}]{sun2020natural}
Lichao Sun. 2020.
\newblock \href {https://arxiv.org/abs/2006.16176} {Natural backdoor attack on
  text data}.
\newblock \emph{ArXiv preprint}, abs/2006.16176.

\bibitem[{Wang et~al.(2022)Wang, Bao, Zhang, and Zhao}]{wang2022rethinking}
Jiayi Wang, Rongzhou Bao, Zhuosheng Zhang, and Hai Zhao. 2022.
\newblock Rethinking textual adversarial defense for pre-trained language
  models.
\newblock \emph{IEEE/ACM Transactions on Audio, Speech, and Language
  Processing}, 30:2526--2540.

\bibitem[{Warstadt et~al.(2019)Warstadt, Singh, and
  Bowman}]{warstadt2019neural}
Alex Warstadt, Amanpreet Singh, and Samuel~R. Bowman. 2019.
\newblock \href {https://doi.org/10.1162/tacl_a_00290} {Neural network
  acceptability judgments}.
\newblock \emph{Transactions of the Association for Computational Linguistics},
  7:625--641.

\bibitem[{Wolf et~al.(2020)Wolf, Debut, Sanh, Chaumond, Delangue, Moi, Cistac,
  Rault, Louf, Funtowicz, Davison, Shleifer, von Platen, Ma, Jernite, Plu, Xu,
  Le~Scao, Gugger, Drame, Lhoest, and Rush}]{wolf2020transformers}
Thomas Wolf, Lysandre Debut, Victor Sanh, Julien Chaumond, Clement Delangue,
  Anthony Moi, Pierric Cistac, Tim Rault, Remi Louf, Morgan Funtowicz, Joe
  Davison, Sam Shleifer, Patrick von Platen, Clara Ma, Yacine Jernite, Julien
  Plu, Canwen Xu, Teven Le~Scao, Sylvain Gugger, Mariama Drame, Quentin Lhoest,
  and Alexander Rush. 2020.
\newblock \href {https://doi.org/10.18653/v1/2020.emnlp-demos.6} {Transformers:
  State-of-the-art natural language processing}.
\newblock In \emph{Proceedings of the 2020 Conference on Empirical Methods in
  Natural Language Processing: System Demonstrations}, pages 38--45, Online.
  Association for Computational Linguistics.

\bibitem[{Wu et~al.(2022)Wu, Gardner, Stenetorp, and Dasigi}]{wu2022generating}
Yuxiang Wu, Matt Gardner, Pontus Stenetorp, and Pradeep Dasigi. 2022.
\newblock \href {https://doi.org/10.18653/v1/2022.acl-long.190} {Generating
  data to mitigate spurious correlations in natural language inference
  datasets}.
\newblock In \emph{Proceedings of the 60th Annual Meeting of the Association
  for Computational Linguistics (Volume 1: Long Papers)}, pages 2660--2676,
  Dublin, Ireland. Association for Computational Linguistics.

\bibitem[{Xu et~al.(2021)Xu, Wang, Li, Borisov, Gunter, and
  Li}]{xu2021detecting}
Xiaojun Xu, Qi~Wang, Huichen Li, Nikita Borisov, Carl~A Gunter, and Bo~Li.
  2021.
\newblock Detecting ai trojans using meta neural analysis.
\newblock In \emph{2021 IEEE Symposium on Security and Privacy (SP)}, pages
  103--120. IEEE.

\bibitem[{Yang et~al.(2021{\natexlab{a}})Yang, Li, Zhang, Ren, Sun, and
  He}]{yang2021careful}
Wenkai Yang, Lei Li, Zhiyuan Zhang, Xuancheng Ren, Xu~Sun, and Bin He.
  2021{\natexlab{a}}.
\newblock \href {https://doi.org/10.18653/v1/2021.naacl-main.165} {Be careful
  about poisoned word embeddings: Exploring the vulnerability of the embedding
  layers in {NLP} models}.
\newblock In \emph{Proceedings of the 2021 Conference of the North American
  Chapter of the Association for Computational Linguistics: Human Language
  Technologies}, pages 2048--2058, Online. Association for Computational
  Linguistics.

\bibitem[{Yang et~al.(2021{\natexlab{b}})Yang, Lin, Li, Zhou, and
  Sun}]{yang2021rap}
Wenkai Yang, Yankai Lin, Peng Li, Jie Zhou, and Xu~Sun. 2021{\natexlab{b}}.
\newblock \href {https://doi.org/10.18653/v1/2021.emnlp-main.659} {{RAP}:
  {R}obustness-{A}ware {P}erturbations for defending against backdoor attacks
  on {NLP} models}.
\newblock In \emph{Proceedings of the 2021 Conference on Empirical Methods in
  Natural Language Processing}, pages 8365--8381, Online and Punta Cana,
  Dominican Republic. Association for Computational Linguistics.

\bibitem[{Zhang et~al.(2021)Zhang, Xiao, Li, Lv, Qi, Liu, Wang, Jiang, and
  Sun}]{zhang2021red}
Zhengyan Zhang, Guangxuan Xiao, Yongwei Li, Tian Lv, Fanchao Qi, Zhiyuan Liu,
  Yasheng Wang, Xin Jiang, and Maosong Sun. 2021.
\newblock \href {https://arxiv.org/abs/2101.06969} {Red alarm for pre-trained
  models: Universal vulnerability to neuron-level backdoor attacks}.
\newblock \emph{ArXiv preprint}, abs/2101.06969.

\bibitem[{Zhu et~al.(2022)Zhu, Qin, Cui, Chen, Zhao, Fu, Deng, Liu, Wang, Wu
  et~al.}]{zhu2022moderate}
Biru Zhu, Yujia Qin, Ganqu Cui, Yangyi Chen, Weilin Zhao, Chong Fu, Yangdong
  Deng, Zhiyuan Liu, Jingang Wang, Wei Wu, et~al. 2022.
\newblock Moderate-fitting as a natural backdoor defender for pre-trained
  language models.
\newblock \emph{Advances in Neural Information Processing Systems},
  35:1086--1099.

\end{thebibliography}
